\documentclass[journal,hidelinks]{IEEEtran}
\IEEEoverridecommandlockouts

\usepackage{cite}
\usepackage{amsmath,amssymb,amsfonts}
\usepackage{algorithmic}
\usepackage{graphicx}
\usepackage{textcomp}
\usepackage{xcolor}
\usepackage{siunitx}

\ifCLASSOPTIONcompsoc
    \usepackage[caption=false,font=normalsize,labelfont=sf,textfont=sf]{subfig}
\else
    \usepackage[caption=false,font=footnotesize]{subfig}
\fi

\newcommand{\supercite}[1]{\textsuperscript{\cite{#1}}}
    
\usepackage{hyperref}
    
\begin{document}

\title{MIMo: A Multi-Modal Infant Model for Studying Cognitive Development}

\author{Dominik Mattern,
Pierre Schumacher,
Francisco M. L\'{o}pez,
Marcel C. Raabe,\\
Markus R. Ernst,
Arthur Aubret,
Jochen Triesch
\thanks{D. Mattern is with the Department of Computer Science and Mathematics, Goethe-University Frankfurt, Frankfurt am Main, Germany.}
\thanks{P. Schumacher is with the Max Planck Institute for Intelligent Systems, Tübingen, Germany and the Hertie Institute for Clinical Brain Research, Tübingen, Germany.}
\thanks{F.M. L\'{o}pez, M. Raabe, M.R. Ernst, A. Aubret and J. Triesch are with the Frankfurt Institute for Advanced Studies, Frankfurt am Main, Germany.}
\thanks{This research was supported by ``The Adaptive Mind'' and ``The Third Wave of Artificial Intelligence'' funded by the Excellence Program of the Hessian Ministry of Higher Education, Science, Research and Art. J. Triesch was supported by the Johanna Quandt foundation.
P. Schumacher was supported by the International Max Planck Research School for Intelligent Systems (IMPRS-IS).}
\thanks{This version of MIMo extends a previous version published at ICDL\cite{mattern2022}.}
}


\maketitle

\begin{abstract}
Human intelligence and human consciousness emerge gradually during the process of cognitive development. Understanding this development is an essential aspect of understanding the human mind and may facilitate the construction of artificial minds with similar properties. Importantly, human cognitive development relies on embodied interactions with the physical and social environment, which is perceived via complementary sensory modalities. These interactions allow the developing mind to probe the causal structure of the world. This is in stark contrast to common machine learning approaches, e.g., for large language models, which are merely passively ``digesting'' large amounts of training data, but are not in control of their sensory inputs. However, computational modeling of the kind of self-determined embodied interactions that lead to human intelligence and consciousness is a formidable challenge. Here we present MIMo, an open-source multi-modal infant model for studying early cognitive development through computer simulations.
MIMo's body is modeled after an 18-month-old child with detailed five-fingered hands. MIMo perceives its surroundings via binocular vision, a vestibular system, proprioception, and touch perception through a full-body virtual skin, while two different actuation models allow control of his body.
We describe the design and interfaces of MIMo and provide examples illustrating its use.
All code is available at \url{https://github.com/trieschlab/MIMo}.

\end{abstract}


\bstctlcite{IEEEexample:BSTcontrol}

\begin{figure}[tbp]
\centering
\subfloat[Full body view]{\includegraphics[width=0.54\linewidth]{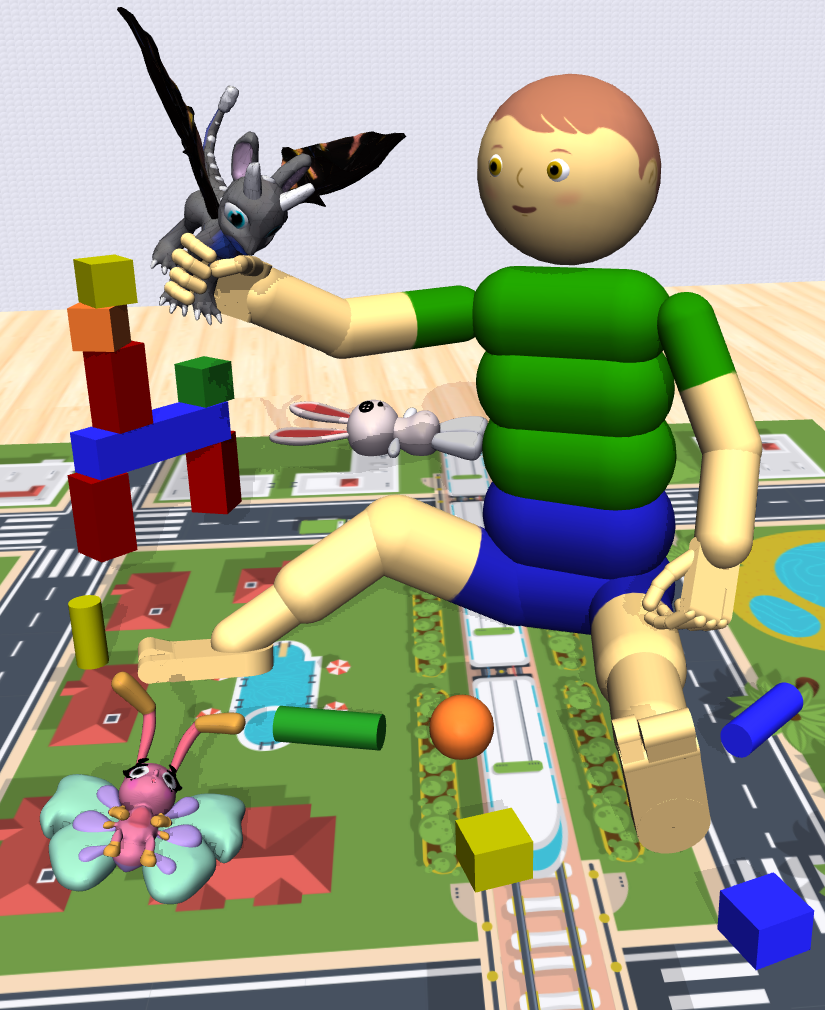}}
\hfil
\subfloat[Facial expressions]{\includegraphics[width=0.37\linewidth]{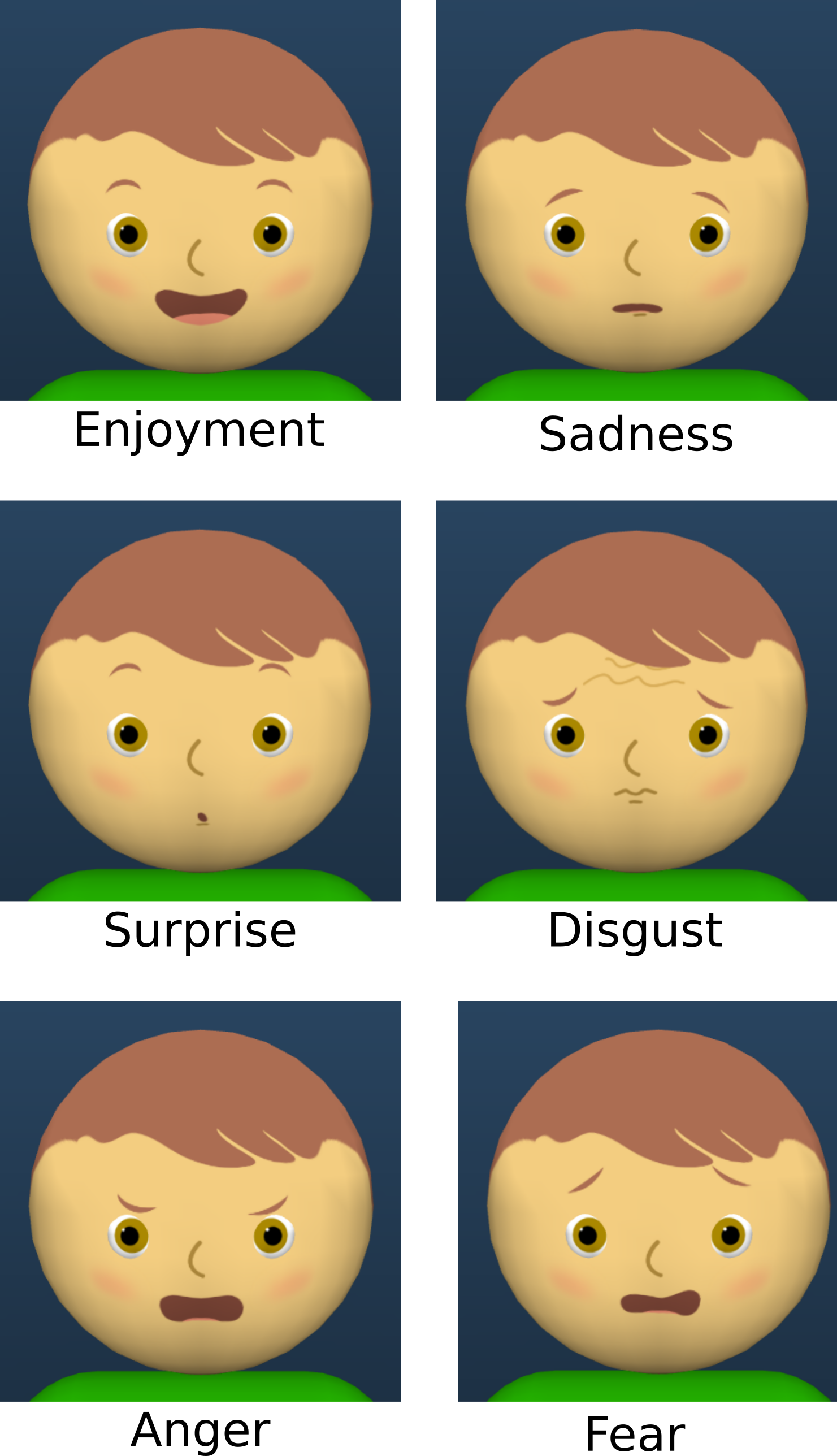}}
\caption{MIMo, the multimodal infant model. (a) MIMo sitting in a room with toys. (b) Six facial expressions of MIMo.}
\label{fig-mimo}
\end{figure}

\section{Introduction}

A good measure of our understanding of a complex system or process is our ability to rebuild it. In the context of human cognitive development this translates to constructing models of how the developing brain comes to control the developing body in increasingly sophisticated ways. Human cognitive development depends crucially on embodied interactions with the physical and social environment, that is sensed through our different sensory modalities. Faithful models of cognitive development therefore need to reproduce these interactions and how they give rise to sensory representations, motor skills, conceptual structures, and a broad range of cognitive abilities.

Importantly, these interactions are largely initiated and controlled by the developing child. For example, an infant controls its visual, proprioceptive, and haptic inputs through its eye and body movements. This ability to control and ``experiment'' with different movements to observe their effects on sensory inputs may be crucial for cognitive development, giving the developing mind a means to probe the causal structure of the world, rather than merely passively observing correlations among sensed variables.

Importantly, such an active, self-controlled form of learning is in stark contrast to a large body of work in Artificial Intelligence (AI), including large language or vision models, which learn certain aspects of the statistical structure of large training datasets without having any control over their inputs. Despite the impressive successes of such models, they may ultimately be severely limited in their ability to understand the causal mechanisms underlying their training data, which will limit their ability to generalize in novel situations.

This should come as no surprise. Every scientist learns that mere correlation is not sufficient evidence for inferring causation. Therefore, scientists routinely exploit the ability to manipulate a system under study using clever experimental designs where they interfere with some variables and observe the effects on others to distill the causal mechanisms at work.

Therefore, recreating the physical interaction with the environment must be a central aspect of rebuilding cognitive development and it may be equally essential for achieving human-like intelligence and consciousness in AIs. The idea that an AI could learn like a developing child can be traced back all the way to Turing \cite{turing1950mind}. However, serious ``Developmental Robotics'' or ``Developmental AI'' efforts have become more common and practical only in the last 20 years, for reviews see \cite{asada2001cognitive,lungarella2003developmental,schmidhuber2006developmental,asada2009cognitive,cangelosi2018babies,doya2019toward}.

There are two options for modeling the interaction of a developing mind with its physical and social environment. The first option is using humanoid robots. This is the approach taken by the Developmental Robotics community. The main advantage is the inherent realism: the system operates in the real world governed by the actual laws of physics. However, working with humanoid robots is expensive, time-consuming, and suffers from the brittleness of today's humanoid hardware. All these factors negatively impact the reproducibility of the research. Furthermore, the sensing abilities of today's robots are usually not comparable to those of actual humans. This is particularly problematic for the sense of touch. The human body is covered by a flexible skin containing various types of mechanoreceptors, thermoreceptors, and nociceptors (pain receptors). These allow us to sense touch, pressure, vibration, temperature, and pain. Today, reproducing such a human-like skin in humanoid robots is still out of reach.

The second option for modeling the interaction of the developing mind with its physical and social environment is to do it completely {\em in silico}. Many physics simulators or game engines are available today that can approximate the physics of such interactions \cite{gan2020threedworld,rohmer2013v,coumans2016pybullet}, for review see \cite{collins2021review}. Among the disadvantages of such an approach are 1) inaccuracies of such simulations due to inevitable approximations and 2) the high computational costs, especially when non-rigid body parts and objects are considered. Nevertheless, the {\em in silico} approach avoids all the problems of working with humanoid robot hardware mentioned above. Furthermore, it ensures perfect reproducibility. Lastly, if simulations can be run (much) faster than real-time, this greatly facilitates the simulation of developmental processes unfolding over long periods of time (weeks, months, or even years).

To support such {\em in silico} research, we here present the open source software platform MIMo, the \underline{M}ulti-Modal \underline{I}nfant \underline{Mo}del (Fig.~\ref{fig-mimo}). MIMo is intended to support two kinds of research: 1) developing computational models of human cognitive development and 2) building developmental AIs that develop more human-like intelligence and consciousness by similarly exploiting their ability to probe the causal structure of the world through their actions.

We have decided to model the body of MIMo after an average 18-month-old child. In total, MIMo has 82 degrees of freedom of the body and 6 degrees of freedom of the eyes. MIMo also features different facial expressions that can be used for studies of social development. To simulate MIMo's interaction with the physical environment we use the MuJoCo physics engine \cite{todorov2012mujoco}, because of its strength at simulating contact physics with friction. Generally, we have aimed for a balance between realism and computational efficiency in the design of MIMo. To accelerate the simulation of physics and touch sensation, MIMo's body is composed of simple rigid shape primitives such as a sphere for the head and capsules for most other body parts. Presently, MIMo features four sensory modalities: binocular vision, proprioception, full-body touch sensation, and a vestibular system. We also plan to add audition in the future.

The remainder of this article describes the detailed design of MIMo and illustrates how to use it. In particular, we present four scenarios where MIMo learns to 1) reach for an object, 2) stand up, 3) touch different locations on his body, and 4) catch a falling ball with his five-fingered hand. These examples are used to illustrate and benchmark potential uses of MIMo. They are not intended as faithful models of how infants acquire these behaviors.

\begin{figure}[b!]
    \centering
    \includegraphics[width=.9\linewidth]{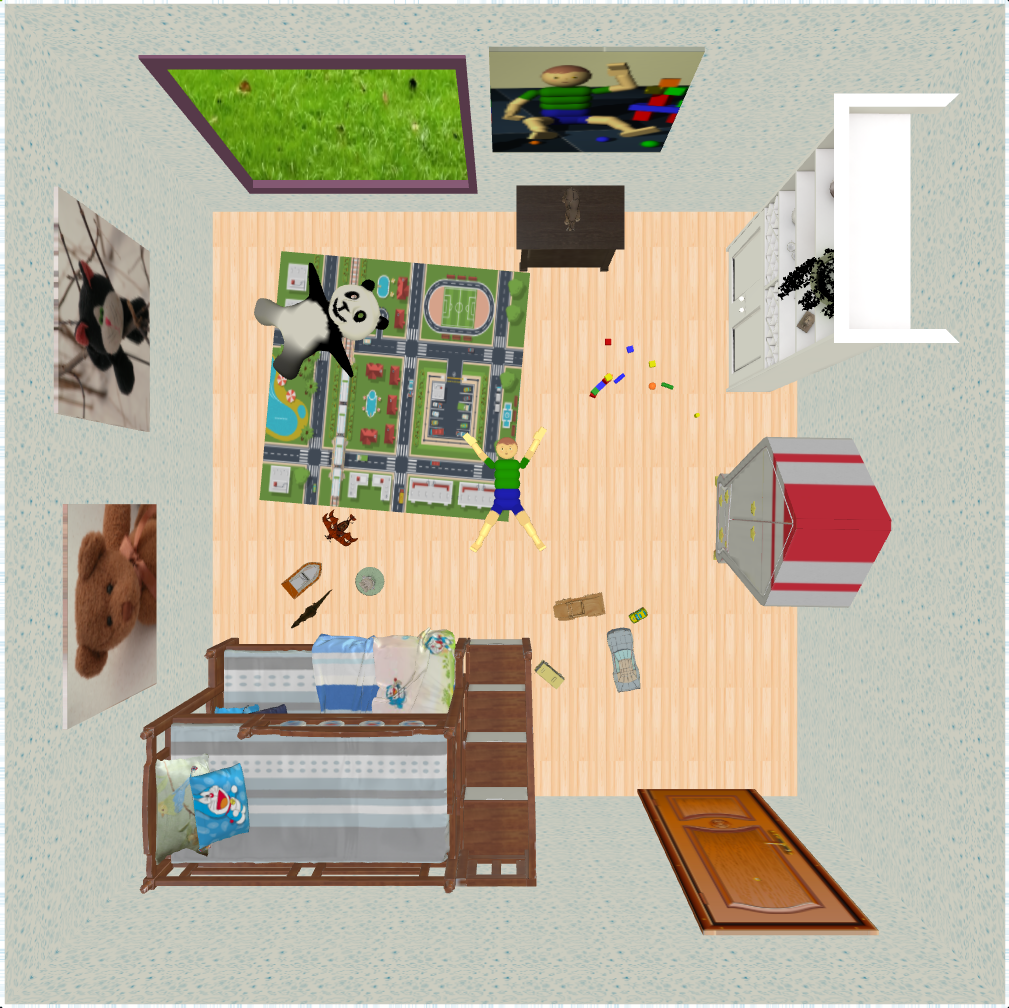}
    \caption{Top view of the play room environment. The room has a size of \(4 \times 4 \, \si{\square\metre} \) and is filled with furniture and toys. Furniture models are taken from the 3D-FUTURE dataset \cite{fu20213d} and toys from the Toys4K 3D Object dataset \cite{Stojanov_2021_CVPR}. Pictures taken from the \textit{Man-made} category of the McGill Calibrated Color Image Database\cite{olmos2004} are placed on the walls. Additionally, the room has a door and a window, which provides a view into a garden.}
    \label{fig:nursery}
\end{figure}

This paper is an extended version of preliminary work previously published at the ICDL 2022 conference \cite{mattern2022}. Compared to the preliminary MIMo version we have added 1) five-fingered hands, 2) a new actuation model that more accurately models the force-generating behaviour and compliance of real muscles, 3) a detailed play room (see Fig. \ref{fig:nursery}), in which MIMo can interact, and 4) a new demo environment of learning to catch a falling ball. A fifth contribution is that we have improved computational efficiency and benchmarks have been updated accordingly.

\section{Related Work}
\label{sec-related-work}

We focus on two major classes of software platforms for simulating cognitive development during embodied interactions with the environment. The first kind is designed to simulate a particular physical robot used in Developmental Robotics research and intended to complement the work with that physical robot. Examples are the iCub simulator \cite{tikhanoff2008open} and the simulator for the NICO robot \cite{kerzel2017nico} that has been implemented using the V-Rep robotics simulation environment \cite{rohmer2013v}. Such platforms typically aim to faithfully reproduce the design and behavior of the physical robot, permitting to substitute work with the real robot through simulations. However, there always remains a notorious gap between simulation and real world. Furthermore, such simulation platforms also inherit any shortcomings of the robot design relative to the human body and human sensing capabilities. For example, if the robot possesses only poor touch sensation, its simulated counterpart will suffer from the same limitation.

The second kind of platform emulates human body and sensing abilities directly and thus is not restricted by limitations of current robotics technology in general or that of specific robots in particular. An early example is the seminal work by Kuniyoshi and Sangawa \cite{kuniyoshi2006early}. More frequently, simulation models of specific aspects of sensorimotor development have been proposed. These typically encompass only a small subset of degrees of freedom and sensory modalities. An early example is work on the development of grasping by Oztop and colleagues \cite{oztop2004infant}. A more recent example is the OpenEyeSim simulator, which has been designed to support modeling the development of active binocular vision \cite{priamikov2016openeyesim} and is built using the OpenSim software for simulating neuromusculoskeletal systems \cite{delp2007opensim}. While widely used in Reinforcement Learning (RL) for locomotion tasks \cite{haarnoja2018sac}, standard humanoids introduced within the MuJoCo \cite{tassa2012synthesis} or Bullet \cite{coumans2016pybullet} platforms only incorporate very limited haptic abilities and do not model a child-like physical  appearance. This may be too limiting, because the structure and physiology of the body constrains the kinds of interactions that are possible.

\section{Physical Design}
\label{sec-physical-design}

MIMo's design is based on the MuJoCo humanoid, consisting of geometric primitives, primarily capsules. His overall body dimensions and proportions were adapted from anthropometric measurements of 16--19 month old infants \cite{anthrokids}, treating the unit of distance in MuJoCo as one meter. 
Body dimensions for which no direct measurement is found in \cite{anthrokids} where induced from other measurements. For example, the length of the lower arm is derived from the elbow-to-fingertip and hand length measurements.
We then made many small adjustments to MIMo's proportions to give him a more natural look, while staying close to the experimental measurements. For example, the upper arms appeared very thin compared to the torso and had their circumferences increased by \SI{0.4}{\centi\metre}, well within the \SI{1.3}{\centi\metre} standard deviation reported by \cite{anthrokids}.

\begin{figure}[tbp]
\centering
\subfloat[Mitten hand]{\includegraphics[width=0.48\linewidth]{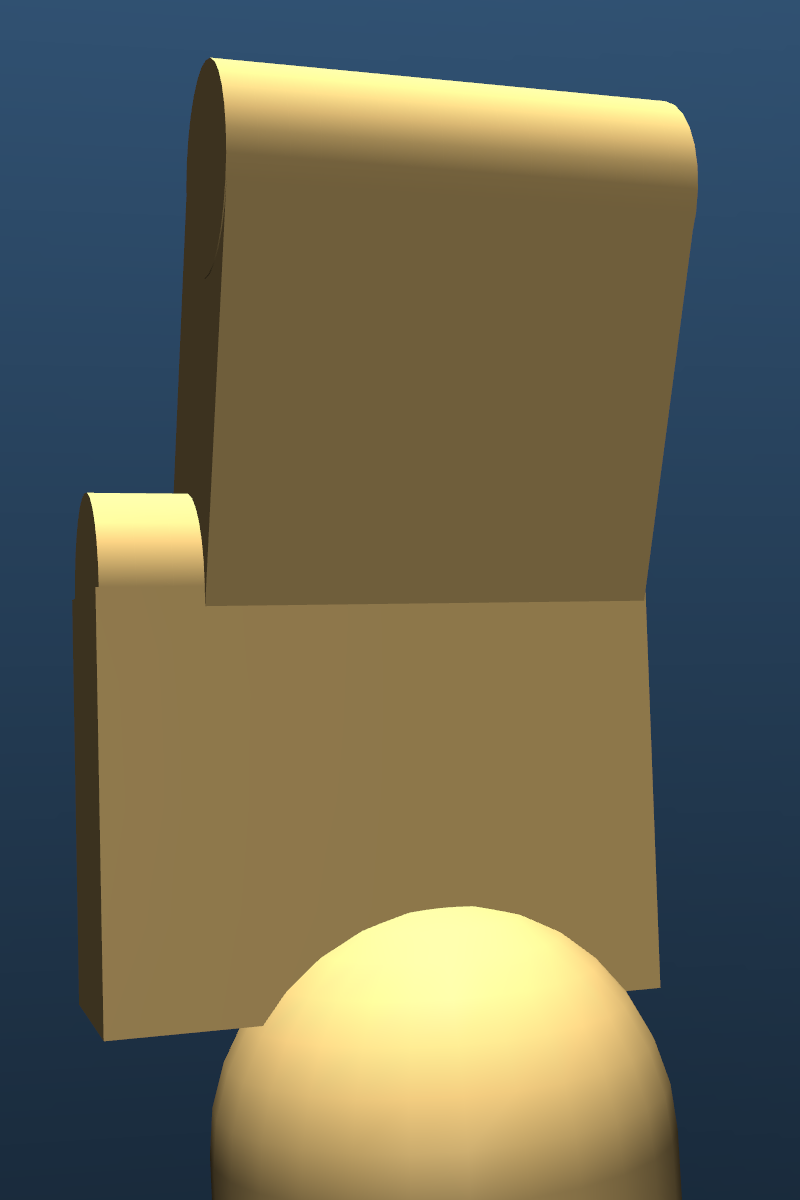}}
\hfil
\subfloat[Full hand]{\includegraphics[width=0.48\linewidth]{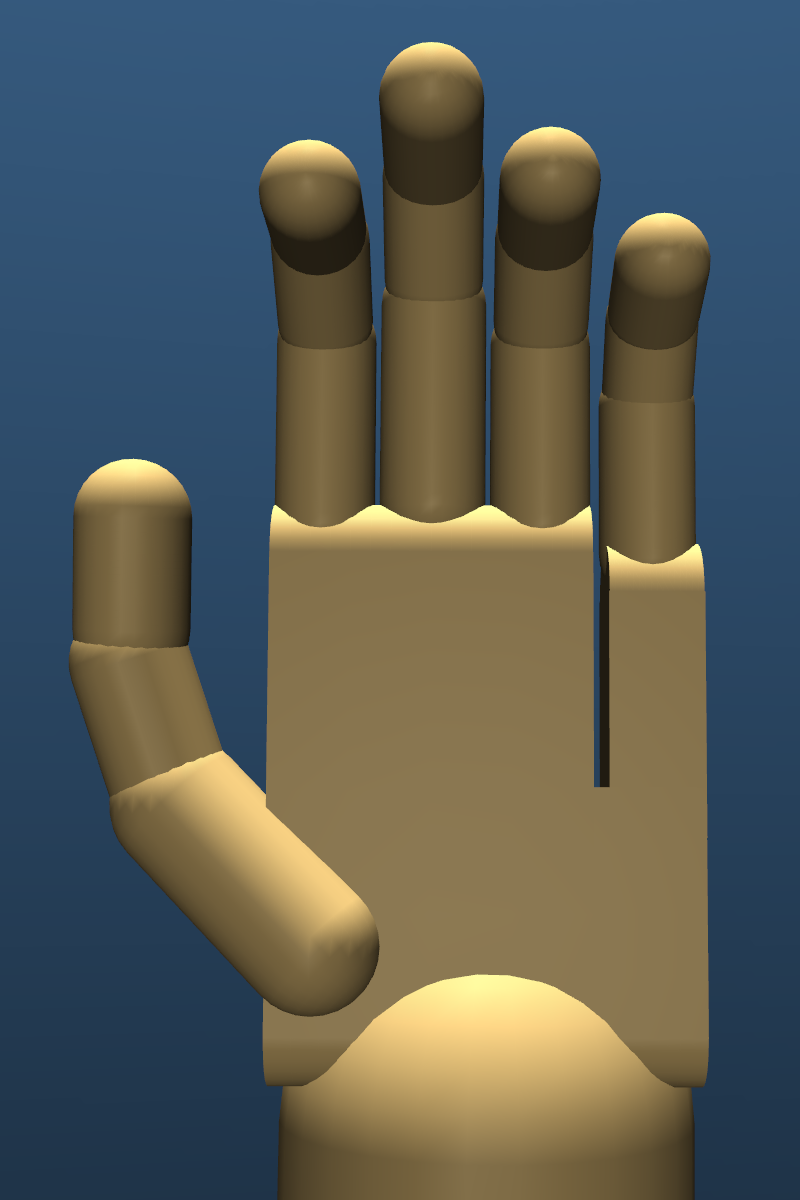}}
\caption{Different versions of MIMo's hands. (a) Simple ``mitten'' hand with 4 degree of freedom. (b) Five-fingered hand with 26 degrees of freedom.}
\label{fig-hands}
\end{figure}

All joints are modeled as a series of one-axis hinges and split into flexion/extension, abduction/adduction or internal/external rotation as appropriate for the joint. For the shoulders we merged the commonly used abduction and flexion axes, instead using horizontal flexion, abduction and internal rotation. This allows MIMo the same total range of motion by combining horizontal flexion and abduction. Keeping both axes would have allowed for an unrealistic, very large total range of motion if both flexion and abduction were at their limit, and MuJoCo's joint limit functions were too simple to prevent this in a satisfactory manner.
We provide two different versions of the hands for MIMo, a simple one with only a single finger and a fully articulated five-fingered version, modified from \cite{kumar2016thesis}, to enable experiments where dexterous manipulation is required (see Fig.~\ref{fig-hands}). In total the mitten hand version has 44 degrees of freedom, while the full hand version has 88. These consist of 30 in the body, 6 in the eyes and 8 and 26 degrees of freedom in the hands for the two versions, respectively.

To produce range of motion and strength measurements, we collected data from a large number of sources \cite{McKay,Eek,Sankar,Gomez,Jordan,Ohman,Watanabe,Gunal,Hughes,Katoh,DaPaz,Bok} and then inferred missing values with the available data. For range of motion we took values from the youngest age reported in the various studies and assumed no change. For muscle strengths we took data from \cite{McKay} as a baseline. These authors consider children aged 3--9 and we used values from the lower end of this range for all joints reported.
Joint strengths missing from \cite{McKay} were induced from other studies by assuming that the relative strengths of joints stay constant, using the strengths of the knee or elbow from \cite{McKay} as reference values. Where required we converted forces to torques using the appropriate lever arms from MIMo based on the methodologies used in the respective source.

Table~\ref{tab:joints} shows the range of motion and strengths for all main joints. We treat extension, adduction and internal rotation as positive and flexion, abduction and external rotation as negative. Citations show the source of the data, while entries without marked sources indicate best guesses based on the other values. For the full hand model we kept the range of motion from \cite{kumar2016thesis}, with finger strengths for individual fingers derived from \cite{ohtsuki1981decrease}. For the mitten hand the strength was derived from \cite{McKay}, with a range of motion that allows him to fully close the hand.

MIMo can also change his facial expression. This is implemented by changing the texture of the head (see Fig. \ref{fig-mimo}). In addition to the neutral expression we provide six extra textures, corresponding to the six basic emotions proposed by \cite{ekman1992emotions} (enjoyment, sadness, surprise, disgust, anger, and fear). These can be used to convey an internal emotional state, for example for studies of social learning with multiple agents.

\begin{table}[tbp]
\caption{Joint range of motion and strength for MIMo.}
\begin{center}
\begin{tabular}{|c|c|c|}
\hline
\textbf{Joint} & \textbf{ROM [°]} & \textbf{Voluntary Torque [Nm]} \\
\hline
Neck flexion/ext. & -70\supercite{McKay} to 80\supercite{McKay} & -1.17\supercite{Jordan}$^*$ to 2.10\supercite{Jordan}$^*$ \\
Neck lateral flex. & -70\supercite{Ohman} to 70\supercite{Ohman} & -1.17 to 1.17 \\
Neck rotation & -111\supercite{Ohman} to 111\supercite{Ohman} & -1.17 to 1.17 \\
& & \\
Trunk flexion/ext. & -61\supercite{Gomez} to 34\supercite{Gomez} & -8.13\supercite{Gomez}$^*$ to 10.58\supercite{Gomez}$^*$ \\
Trunk lateral flex. & -41\supercite{Gomez} to 41\supercite{Gomez} & -7.25\supercite{Gomez}$^*$ to 7.25\supercite{Gomez}$^*$ \\
Trunk rotation & -36\supercite{Gomez} to 36\supercite{Gomez} & -3.63\supercite{Gomez}$^*$ to 3.63\supercite{Gomez}$^*$ \\
& & \\
Shoulder horizontal & -118\supercite{Gunal} to 28\supercite{Gunal} & -1.8\supercite{Katoh}$^*$ to 1.8\supercite{Katoh}$^*$ \\
Shoulder flexion/ext. & -183\supercite{Watanabe} to 84\supercite{Watanabe} & -2.75\supercite{Hughes}$^*$ to 4\supercite{Hughes}$^*$\\
Shoulder rotation & -99\supercite{McKay} to 67\supercite{McKay} & -1.6\supercite{Hughes}$^*$ to 2.5\supercite{Hughes}$^*$ \\
& & \\
Elbow flexion/ext. & -146\supercite{McKay} to 5\supercite{Watanabe} & -3.6\supercite{McKay} to 3.0\supercite{McKay} \\
& & \\
Wrist palmar/dorsi & -92\supercite{Watanabe} to 86\supercite{Watanabe} & -1.24\supercite{vanswearingen1983measuring} to 0.7\supercite{Eek}$^*$ \\
Wrist ulnar/radial & -53\supercite{DaPaz} to 48\supercite{DaPaz} & -0.83\supercite{vanswearingen1983measuring} to 0.95\supercite{vanswearingen1983measuring} \\
Wrist rotation & -90\supercite{Watanabe} to 90\supercite{Watanabe} & -0.7 to 0.7 \\
& & \\
Hip flexion/ext. & -133\supercite{McKay} to 20\supercite{Sankar} & -8\supercite{Eek}$^*$ to 8$^*$\supercite{Eek}$^*$ \\
Hip ab-/adduction & -51\supercite{Sankar} to 17\supercite{Sankar} & -6.24\supercite{Eek}$^*$ to 6.24\supercite{McKay} \\
Hip rotation & -32\supercite{McKay} to 41\supercite{McKay} & -2.66\supercite{McKay}  to  3.54\supercite{McKay} \\
& & \\
Knee flexion/ext. & -145\supercite{McKay} to 4\supercite{McKay} & -6.5\supercite{McKay} to 10\supercite{McKay} \\
& & \\
Ankle plantar/dorsi & -63\supercite{McKay} to 32\supercite{McKay} & -3.78\supercite{McKay} to 1.89\supercite{McKay} \\
Ankle e-/inversion & -33\supercite{Bok} to 31\supercite{Bok} & -1.06\supercite{Bok}$^*$ to 1.16\supercite{Bok}$^*$ \\
Ankle rotation & -20 to 30 & -1.2 to 1.2 \\
\hline
\multicolumn{3}{c}{\parbox{0.94\linewidth}{* Reported value scaled to be proportional to knee or elbow reference.}}
\end{tabular}
\label{tab:joints}
\end{center}
\end{table}

\section{Actuation Models}

We have implemented two actuation models incorporating muscle-like properties and providing different trade-offs regarding accuracy and computational efficiency. In both models we pursue a ``big-picture'' approach, grouping the various muscle groups acting on each joint into 1 actuator per movement axis, with each model using a different internal mechanism for torque generation.

\subsection{Spring-Damper Model}

In the first approach, each actuator is modeled as a combination of a motor with a spring-damper system. This approach requires little run time, while providing reasonable accuracy. The motor acts as the voluntary muscle force while the spring acts to return the joint to its neutral position. The spring and the damper loosely approximate viscoelastic characteristics of real muscles. Note that the spring opposes any motion deviating from the equilibrium position.
The joint strengths were set as the maximum output torque of the motor.
Damper strengths were adjusted manually to ensure simulation stability. 
Spring stiffness for most joints was set such that at maximum joint deflection, net torque is reduced by 10\

\subsection{Muscle Model}
\label{sec-muscles}

The second approach implements a muscle-like model based on \cite{wochner2022learning}, with adjusted parameters for MIMo. It is similar to MuJoCo's muscle actuators, while being more flexible regarding muscle parameters.
Compared to the spring-damper model, this approach more accurately models the behavior of real muscles, allowing in particular for adaptive compliance, at the cost of increased run time (see Sec. \ref{sec-compliance}, \ref{sec-benchmarks}).
Each actuator is modeled as two opposing, independently controllable muscles. The output torque of the whole actuator is the sum of the torques of both muscles. We also retain 5\
The torque of a single muscle depends on its current length $l$, velocity $\dot{l}$ and muscle activity $a$ through: 
\begin{equation}
    \tau = \left( F_{L}(l)\,F_{V}(\dot{l}, v_{\textrm{max}})\,\textcolor{black}{a(u,t)} + F_{P}(l) \right) \textcolor{black}{d} f_{\textrm{max}},
\end{equation}
where $F_L$ is the force-length curve, $F_V$ the force-velocity curve and $F_P$ is a passive elastic force. The factor $d$ accounts for the moment arm. \textcolor{black}{The muscle activity $a$ is dependent on time $t$ and the control input $u$ through
\begin{equation}
    \frac{\text{d}a}{\text{d}t} = \frac{1}{k}(u-a),
\end{equation}
with time constant $k=$ \SI{10}{\milli\second}.} See\cite{wochner2022learning} for more details. 
We adjusted two of the parameters, $f_{\textrm{max}}$ and $v_{\textrm{max}}$, for MIMo specifically. The parameter $f_{\textrm{max}}$ scales the normalized, unit-less force-curves into the proper range, while $v_{\textrm{max}}$ scales the damping properties of the $F_{V}$ curve.

To determine appropriate $f_{\textrm{max}}$ values we recreated the experimental setups from previous studies measuring the joint strengths~(see Sec.~\ref{sec-physical-design}). A given joint is fixed in position and maximum control input are applied. After a short time the applied torque was measured and $f_{\textrm{max}}$ adjusted so that the applied torque matched the values from the literature.

As $v_{\textrm{max}}$ measurements do not exist for most joints, we produce an initial set of values for all joints that produce stable behaviour and then scale them to the appropriate range by a common factor derived from one of the few reference values from the literature.
To create the initial set, we use an iterative approach. We set up a custom environment in which MIMo is suspended in the air with gravity disabled. After setting a starting $v_{\textrm{max}}$ value, we take Bernoulli-distributed control inputs of 0 or 1~(i.e.\ bang-bang-control) every two seconds for 30 seconds. We determine the maximum achieved velocity and use it as the new $v_{\textrm{max}}$. This process is repeated for some number of episodes. To ensure convergence, we update $v_{\textrm{max}}$ using a learning rate~$\alpha\in[0,1]$, which is exponentially decayed over multiple iterations. We also leverage symmetry of the body by averaging between left and right versions of joints.
We then determine a scaling factor between our $v_{\textrm{max}}$ value and the one from \cite{frey2012knee}, both for the knee joint, and apply this factor to our values for all joints. We chose the knee joint for this as we have a reference value for it, and because many of the strength measurements were induced using the same method of proportional scaling, also using the knee joint (see Sec.~\ref{sec-physical-design}).

\subsection{Muscle Compliance}
\label{sec-compliance}

\begin{figure}[tb]
  \centering
  \includegraphics[width=\linewidth]{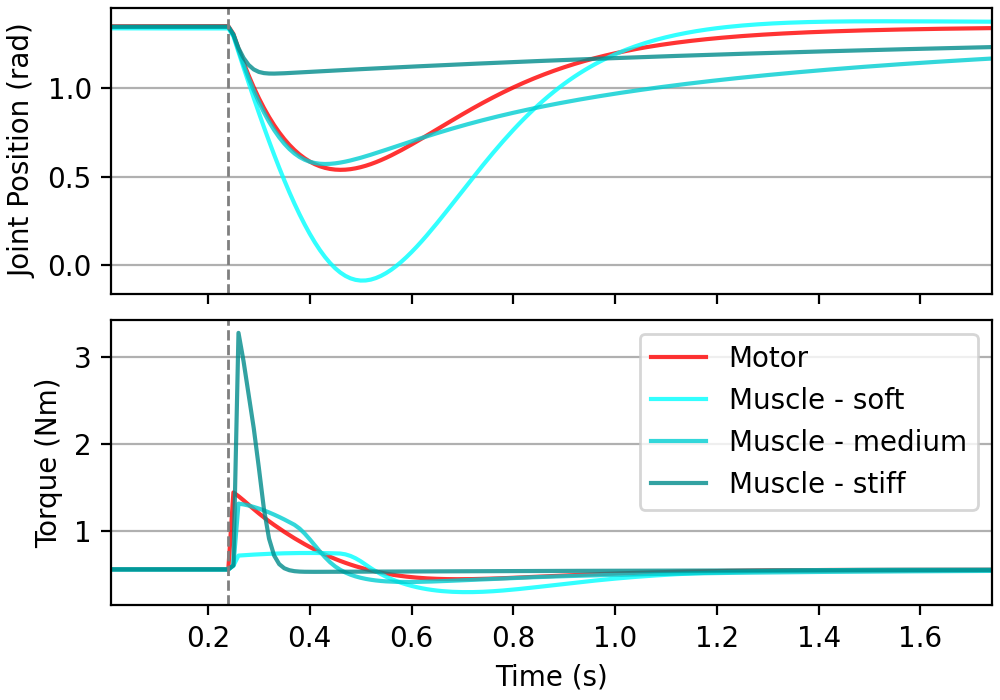}
  \caption{These charts show MIMo's shoulder position and actuator torque as a ball is dropped onto his outstretched arm. The vertical line indicates the moment of impact. For the motor in the spring-damper model, the actual motor torque is constant, but net-actuation torque changes due to the spring-damper as the joint is displaced, with the spring offering less resistance to the motor and the damper acting to slow the motion.}
  \label{fig:compliance}
\end{figure}

A key property of muscular systems is adaptive compliance. When a limb is perturbed from a stable position, the muscles and tendons stretch elastically, altering the force response. It has been shown that these properties support motion stabilization during certain tasks~\cite{dampingmo2020,muscledamping22}. The degree of compliance can be adapted by the relative contraction levels of the muscle pair.

In our spring-damper model, adaptive compliance is not possible, as spring and damper properties are fixed parameters.
The muscle model features two independently controllable muscles for each joint, instead of a single actuator. This allows for the generation of the same movement with different co-contraction levels, naturally altering the level of compliance and the response to perturbations. This effect is present in humans, as seen in \cite{lang1999cerebellar}. Here, two groups of people were asked to catch falling balls while keeping their hands steady. In the healthy group, pre-stiffening of the wrist and elbow muscles was observed before impact, reducing deflection when the ball hit their hand. In the second group, suffering of cerebellar ataxia, the muscle activity did not increase until after impact. This lead to a larger deviation in hand position.

To demonstrate this capability, we created a scenario where MIMo holds his arm outstretched in a stable pose before having a ball dropped on his arm. All joints except the shoulder joint were locked in place. The shoulder joint was held in position by the associated actuator.
We measure shoulder position and actuator torque as the arm deflects from the impact and then returns to position. The control input to the actuator was chosen to hold the arm stable before impact and was constant during the whole simulation. We repeated this for both actuation models.
Control inputs were chosen manually. We picked three sets of inputs with different activation levels for the muscle model. There was only a single input for the spring-damper motors that kept the arm steady.

The result is shown in Fig.~\ref{fig:compliance}. In the muscle model, higher co-contraction levels lead to less deflection as the resisting force increases more strongly after the impact. This adaptive compliance is not possible using the spring-damper model.
Torque curves and overall response are quite similar between the motor and the medium stiffness muscles, demonstrating that the spring-damper model is reasonably accurate for experiments where high muscle fidelity is not critical.

\subsection{Costs}
We implement two versions of an action cost function \textcolor{black}{$c(u)$} commonly used in reinforcement learning. Both use a measure for the amount of effort required for the actuation, but the second version \textcolor{black}{$c_w(u)$} additionally considers the strength of the actuators. 
These functions have the same general form for both actuation models.
For the spring-damper model we have:
\begin{align}
    c(u) & = \frac{\sum_i u_i^2}{n}, \\
    c_w(u) & = \frac{\sum_i u_i^2 T_i}{n\sum_i T_i},
\end{align}
where $n$ is the number of motors in the scene and $u_i$ and $T_i$ are the control input and maximum torque of motor $i$, respectively.
For the muscle model we have
\begin{align}
    c(u) & = \frac{\sum_i \textcolor{black}{a_i(u_i)^2}}{n}, \\
    c_w(u) & = \frac{\sum_i \textcolor{black}{a_i(u_i)^2} f_{\textrm{max}_i}}{n\sum_i f_{\textrm{max}_i}}, \label{eq-musclecost}
\end{align}
where $n$ is the number of muscles, $a_i$ is the activity of muscle $i$ and $f_{\textrm{max}_i}$ is as described in Section~\ref{sec-muscles}.

Both actuation models expose any parameters or intermediate results for computing custom action penalties or metabolic costs.

\section{Multimodal sensing}
\label{sec-multimodal-sensing}

As with the actuation models, we pursue a big-picture approach for our models of the sensory modalities. For example, Golgi tendon organs measure the strain between muscles and their tendons and thus measure the mechanical load on the muscle, while muscle spindles measure muscle contraction and velocity. All these receptors over a single joint in essence measure the joint position, velocity and applied torque over that joint. We do not model all these various receptors themselves, instead computing the quantities directly.

Four different modalities are implemented: Proprioception, vision, a vestibular system, and a full-body touch sensitive skin. The first three have simple implementations and we list the information they collect below. The touch-sensitive skin is described in more detail.

MIMo's proprioceptive system provides him with joint positions and velocities, torques applied across each joint, and limit sensors that \textcolor{black}{activate} as each joint approaches its range of motion limits. In addition there are also quantities from the actuation model, which depend on the specific implementation. For example the muscle model provides the current muscle activation for both muscles in each actuator.

Vision is implemented through two color cameras located in his eyes. The range of motion is $\pm$45° horizontally, -47° to 33° vertically \cite{Lee} and $\pm$8° torsionally \cite{rosenbaum1999strabismus}. The cameras render two RGB images with a 60º field of view, equivalent to the central vision of humans \cite{strasburger2011vision}.

The vestibular system, which provides our sense of balance, is implemented as a combination gyroscope and accelerometer located at the center of MIMo's head.

\subsection{Touch Perception}

Human touch sensation is produced by a variety of receptors responding towards specific aspects of touch, such as the {\em Slowly Adapting type 1} \textcolor{black}{(SA1)}, which responds primarily to direct pressure and coarse texture or the {\em Rapidly Adapting} \textcolor{black}{(RA)} type for slip and fine texture \cite{johnson1992neural}. As with the other modalities, we model these types very simply, ignoring signal travel times and condensing the various types of receptors into a single generic ``touch'' sensor type that measures normal and frictional forces at its location. Sensor points are spread evenly on each individual body part, with the sensor density varying between body parts based on the two-point discrimination distances by \cite{Mancini}. Thus, the front and the back of the palm have the same sensor density, but not the fingertips.

MuJoCo only performs rigid-body simulations in which all contacts are treated as point contacts. Area contacts between flat surfaces produce multiple contact points, for example at the corners of the contact area. As soft-body physics is very computationally expensive, we do not adjust these physics but weakly simulate the deformation of the skin to these point contacts, by distributing the point contact forces over nearby sensor points according to a surface response function. Our response function decreases sensed force linearly with distance from the contact point and then normalizes over all sensor points, such that the total sensed force remains identical to the initial point contact force from MuJoCo. The distance measure is the euclidean distance as geodesic approximations were too computationally expensive. To avoid bleed-through to the opposite side of thin bodies (such as the palm), we only select nearby sensor points with a breadth-first-search on the mesh of sensor points. While also expensive, the results of this search can be cached and reused, leading to a only a small run time penalty compared to using just the euclidean distance.

All of these aspects, from the sensor point density through the response function can be easily adjusted or expanded.

\begin{figure}[tb]
\centering
\subfloat[Binocular vision]{\includegraphics[width=0.45\linewidth]{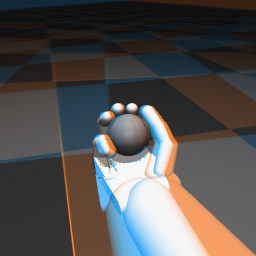}}
\hfil
\subfloat[Touch perception]{\includegraphics[width=0.41\linewidth]{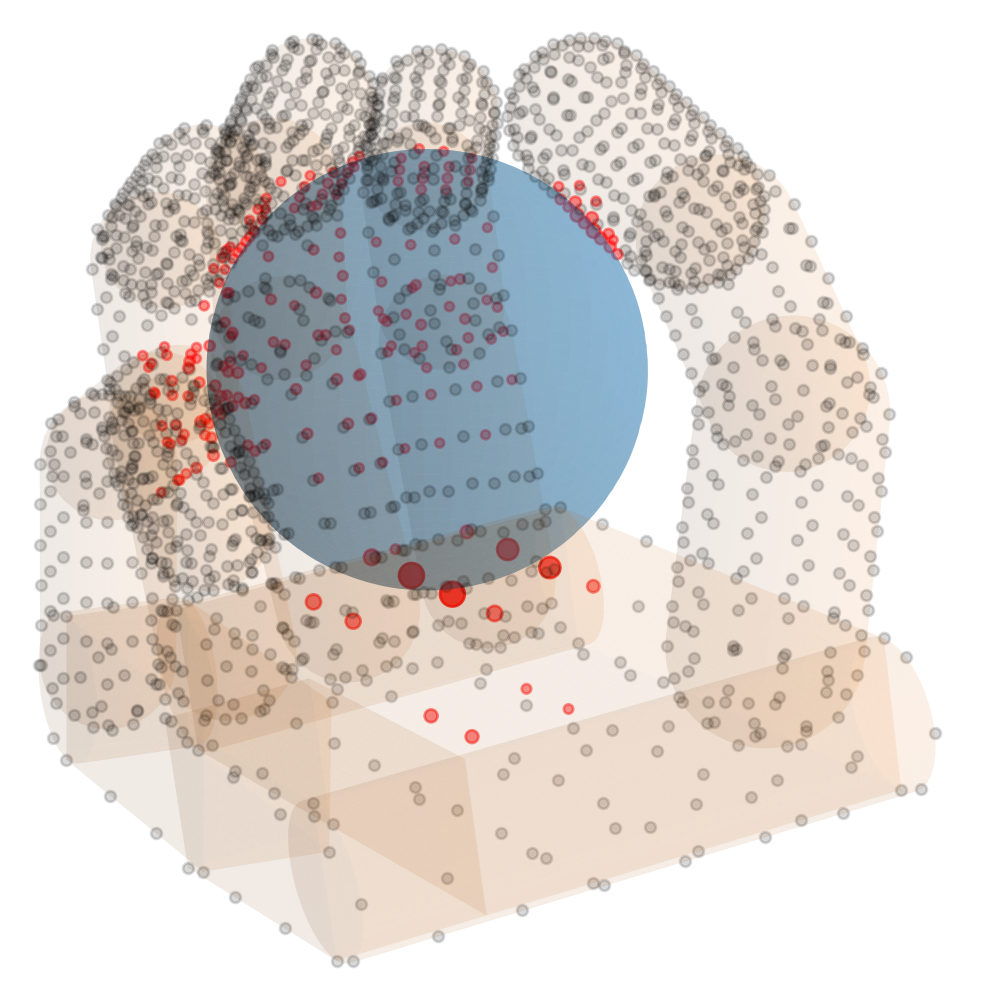}}
\caption{MIMo's multimodal perception while holding a ball. (a) Anaglyph of left and right eye views. (b) Visualization of touch sensors in the hand, with those reporting a force colored red. The size of the circle corresponds to the amount of force sensed.
}
\label{fig-sensors}
\end{figure}

\section{Application Programming Interface}
\label{sec-API}

Our code is written in Python and built as an OpenAI gym \cite{brockman2016gym} environment to allow easy integration into existing experimental setups and take advantage of the large amount of documentation and third-party libraries that already exist, such as the stable baselines library\cite{stable-baselines3}. This environment is intended as an abstract base class that will be subclassed and adapted by other environments for specific experiments.
These subclasses would handle reward structure, sensor limits or any additional constraints. Underdeveloped or limited sensors can be implemented through their configurations, but for the most part the user would implement any perceptual constraints.
The environment is set up to facilitate this in a straightforward way.
The configuration of the sensory modalities, such as the density of the touch sensors or the field of view and resolution of the visual system can be adjusted or disabled easily during initialization without modifying the underlying MuJoCo XMLs.
Swapping between actuation models only requires a single line change as well.

The action and observation spaces are generated automatically based on the configuration of the MuJoCo XMLs and the sensor modules. Disabling touch perception also removes the associated entry from the observation space.
All of the sensory modalities are programmed as separate modules and can be readily attached to any MuJoCo-based gym environment.

\textcolor{black}{Our simulation is time discrete with two different time step types. The physics time step determines the temporal resolution of the physics simulation and choosing a sufficiently small one is important for simulation stability.}

\textcolor{black}{The second step type are control steps, during which observations are collected and the control algorithm is queried for new motor commands. Control frequency is often an important factor for RL algorithms. Control step size must be an integer multiple of the physics step size.}

\section{Experiments}
\label{sec-experiments}

\subsection{Illustrations of learning}

\textcolor{black}{While RL is a powerful framework to generate controllers, its chaotic nature and requirements for large amounts of data require a stable and efficient model. In the following, we demonstrate that MIMo is well-suited for RL even while including multimodal inputs. Note that badly designed models can negatively affect RL performance, even if they are physically accurate in some regimes.} We use four example tasks: reaching for objects, standing up, self-body knowledge and catching a falling ball. In these tasks, two different state-of-the-art, widely used deep RL algorithms, Proximal Policy Optimization (PPO) \cite{schulman2017ppo} and Soft Actor-Critic (SAC) \cite{haarnoja2018sac} \textcolor{black}{are trained to optimize task-dependent reward functions by controlling MiMo's actuators.} \textcolor{black}{While both algorithms are actor-critics, PPO is a widely-used on-policy approach closely related to trust-region algorithms~\cite{pmlr-v37-schulman15} and SAC is an off-policy algorithm with a maximum-entropy formulation~\cite{han2021max}.}

\begin{figure*}[tb]
\centering
\subfloat[Reaching for objects]{
\includegraphics[width=0.33\linewidth]{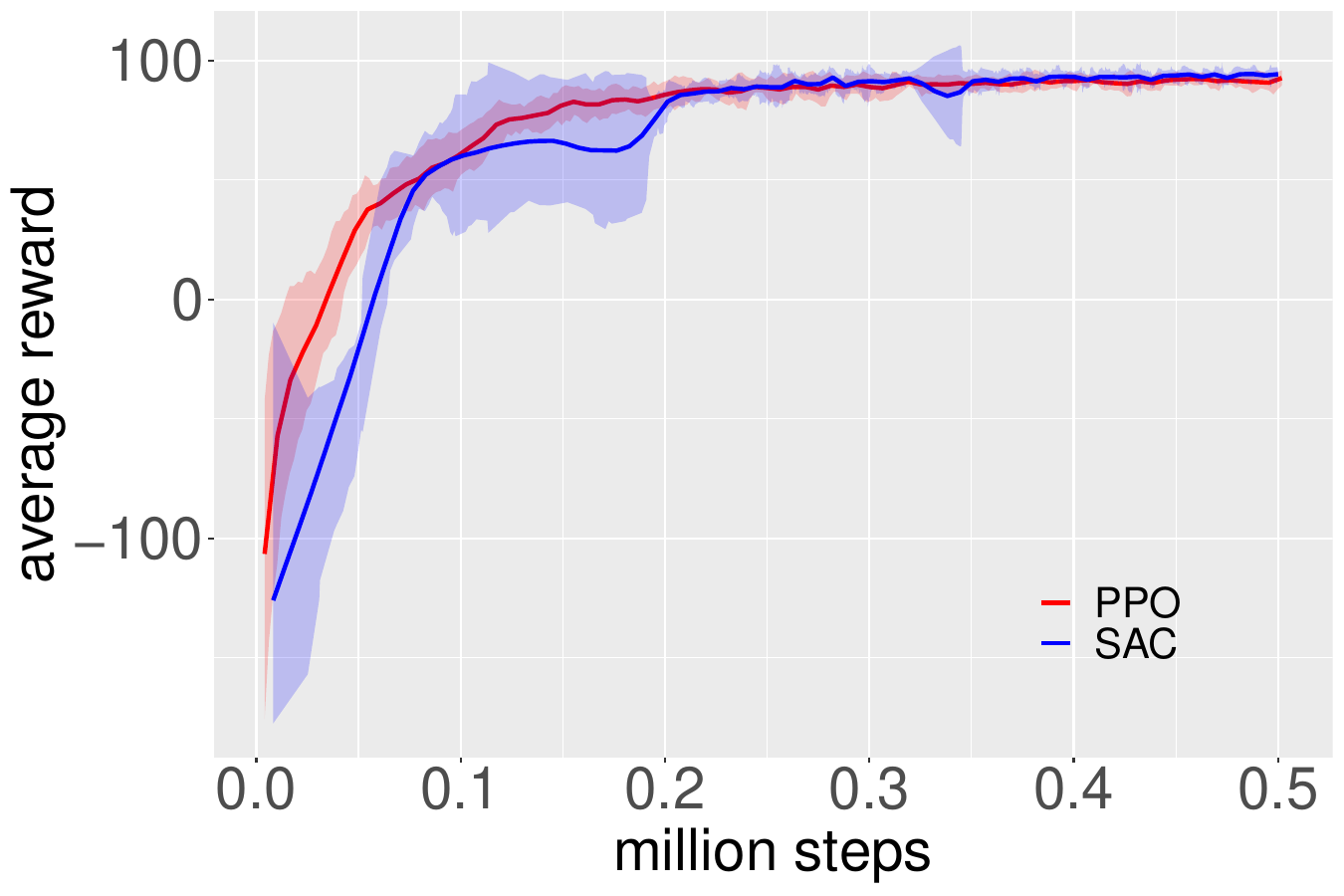}
\includegraphics[width=0.11\linewidth]{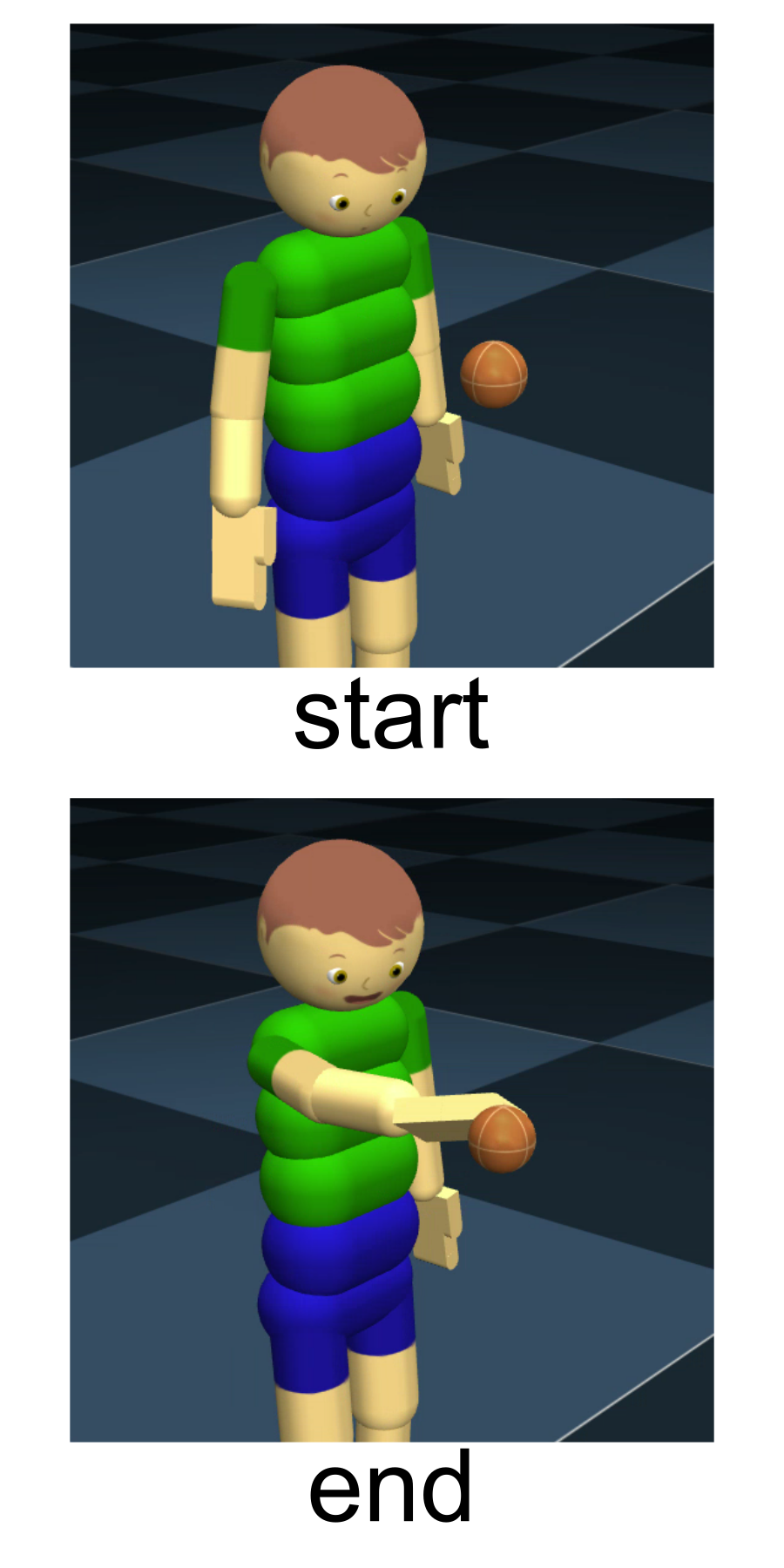}
} 
\hfil
\subfloat[Standing up]{
\includegraphics[width=0.33\linewidth]{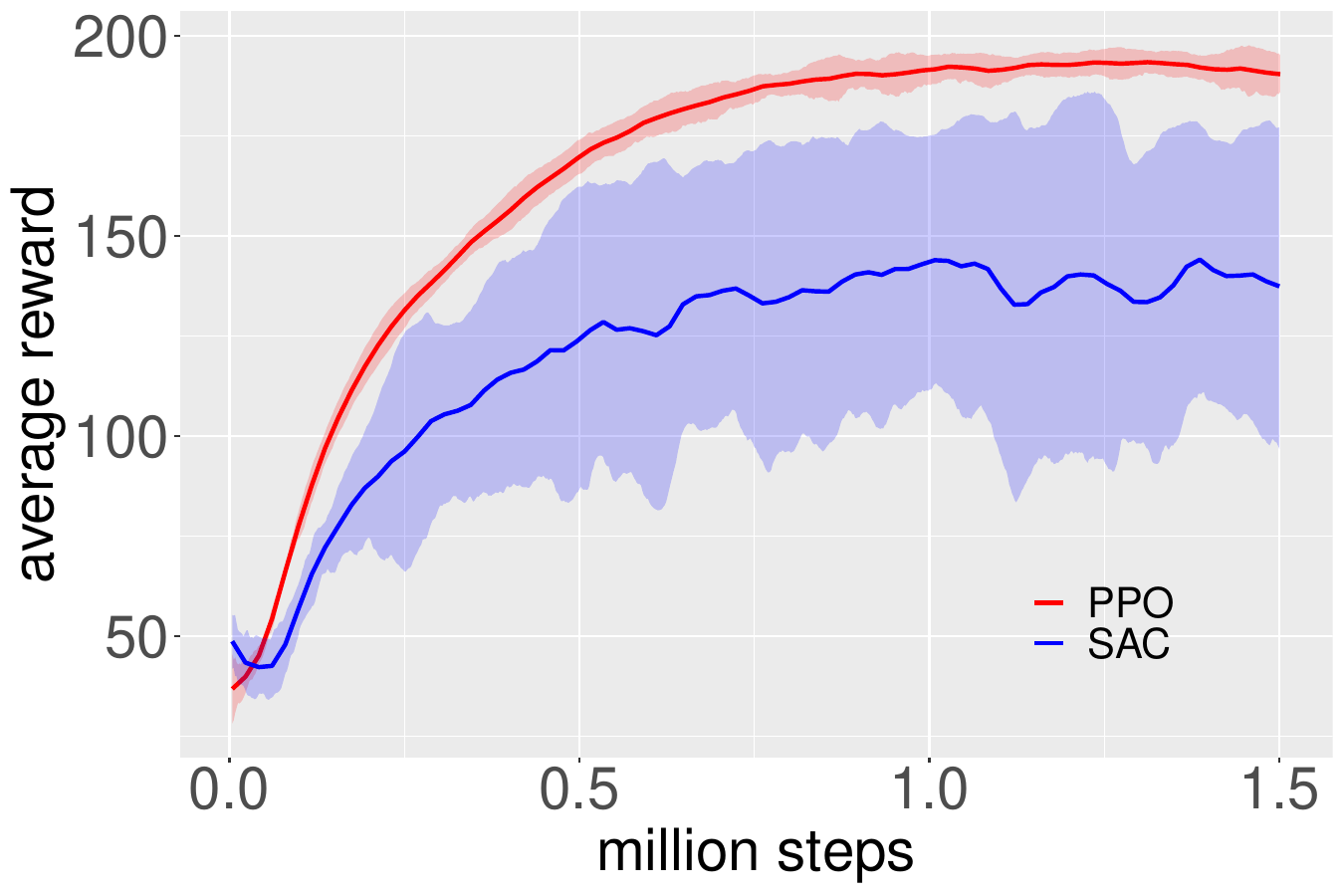}
\includegraphics[width=0.11\linewidth]{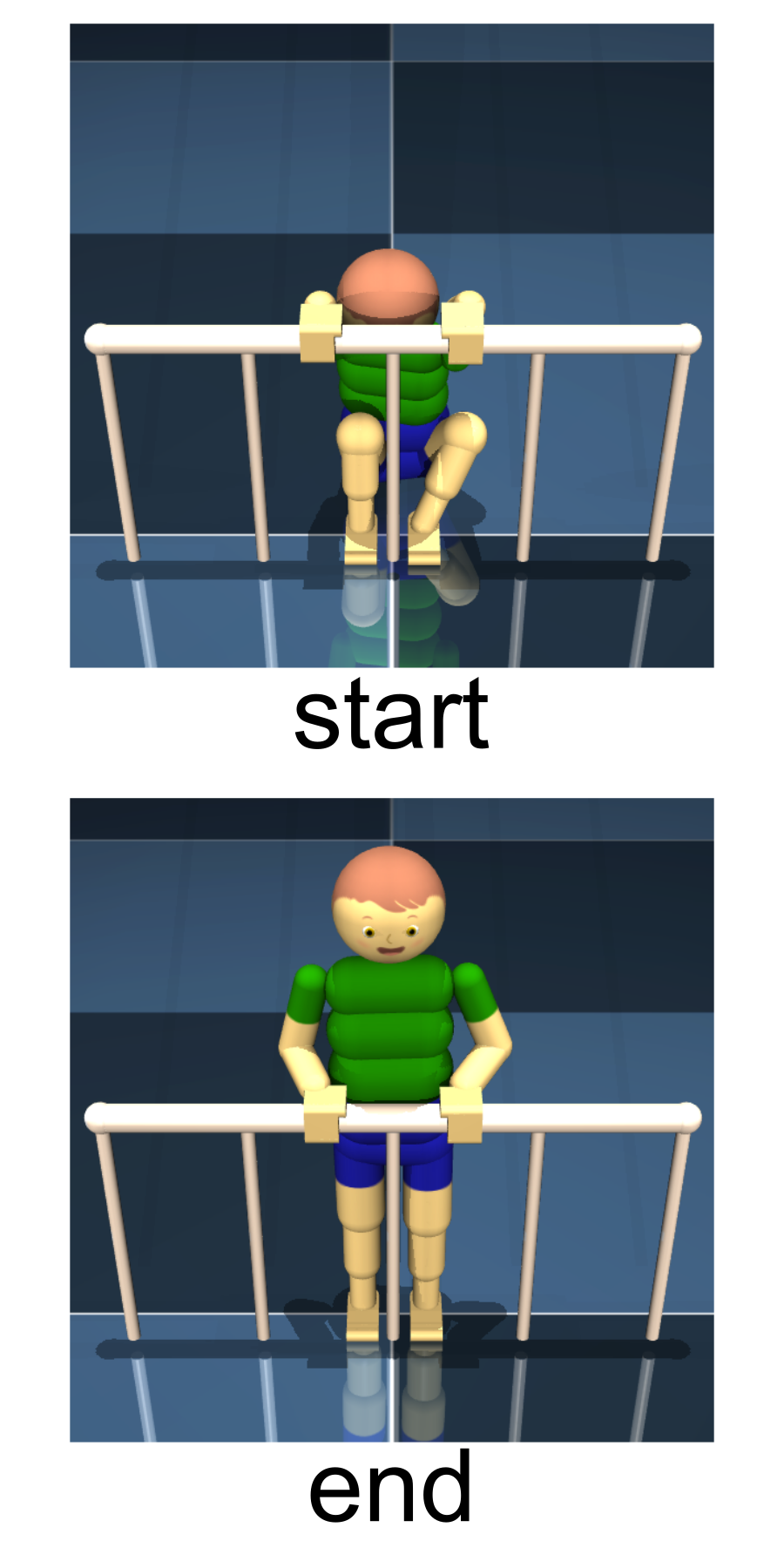}
}
\\
\subfloat[Self-body knowledge]{
\includegraphics[width=0.33\linewidth]{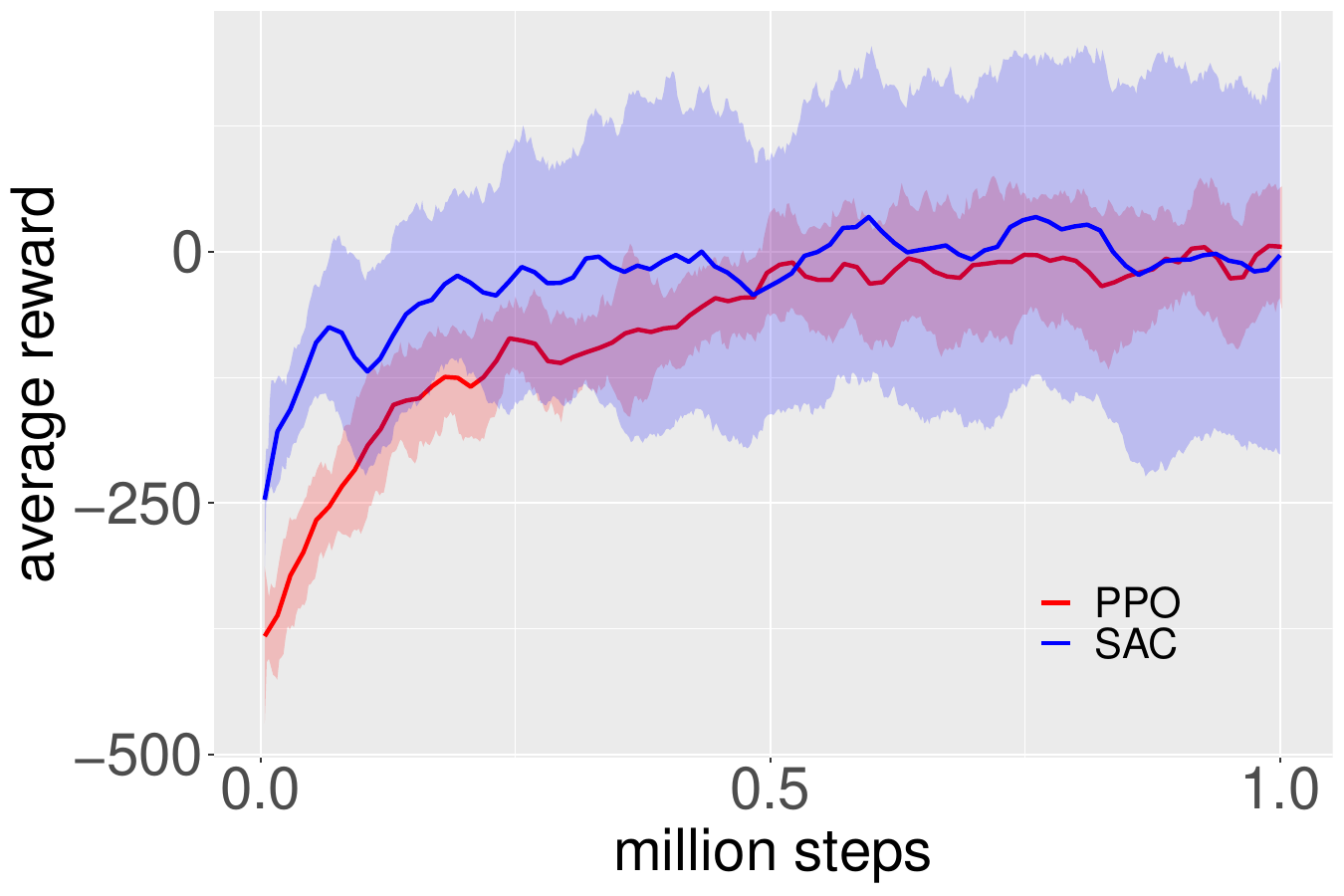}
\includegraphics[width=0.11\linewidth]{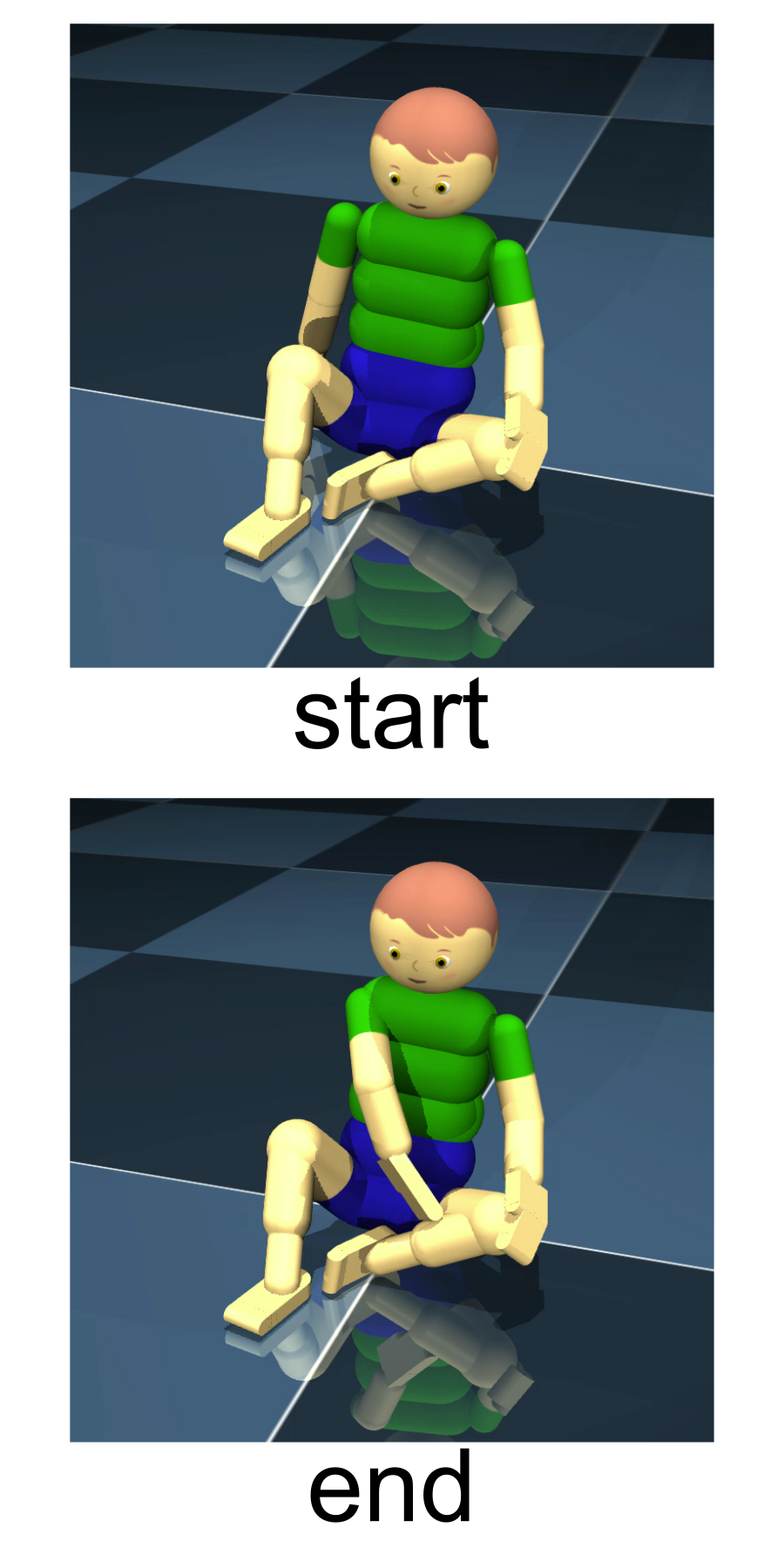}
}
\hfil
\subfloat[Catching objects]{
\includegraphics[width=0.33\linewidth]{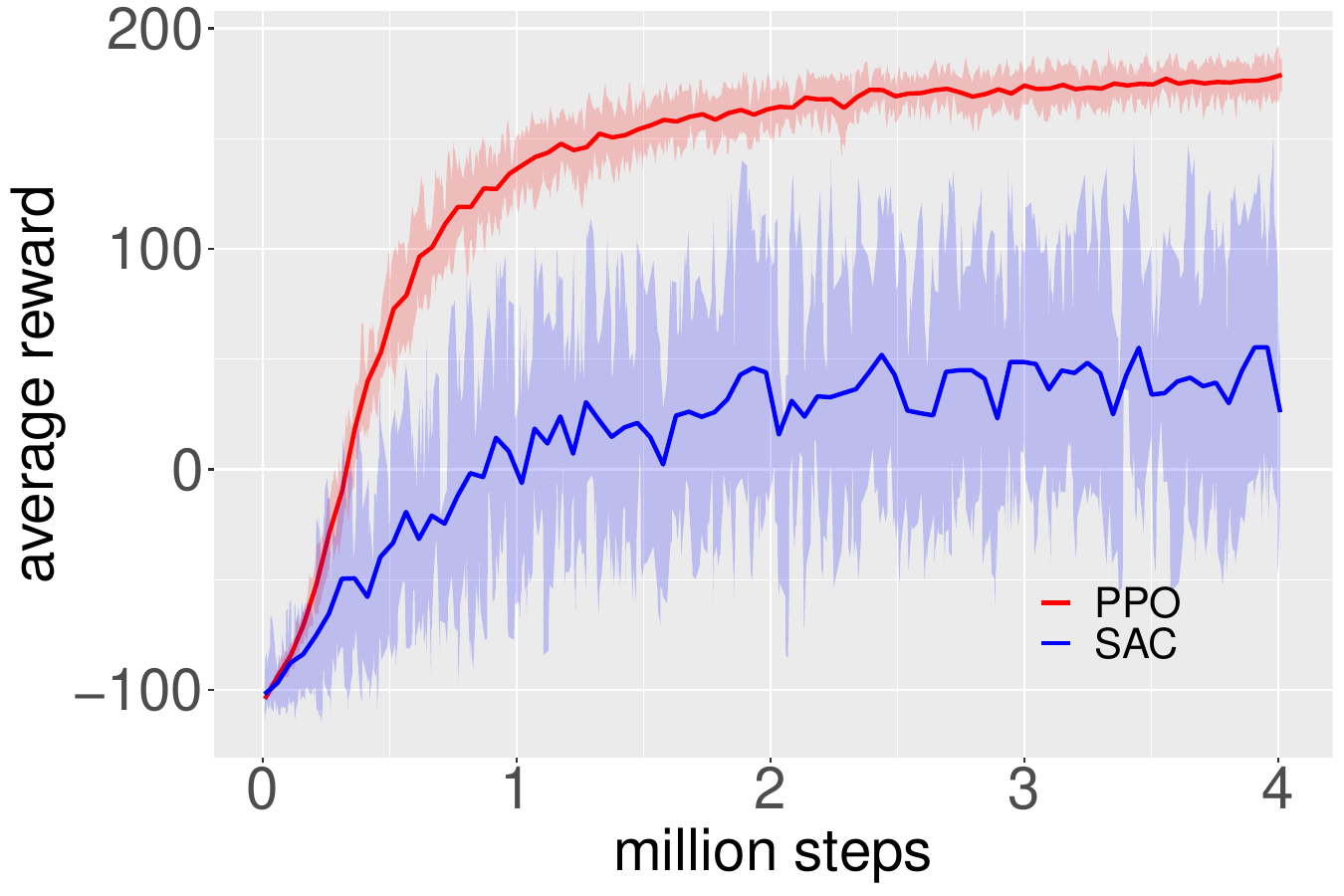}
\includegraphics[width=0.11\linewidth]{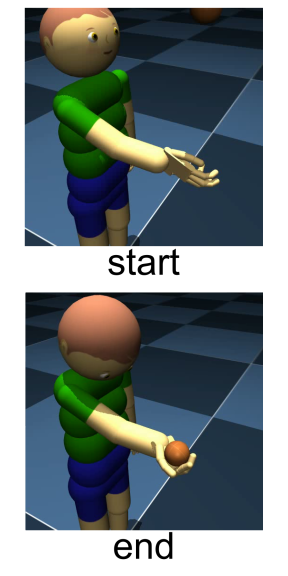}
}
\caption{Comparison of learning curves for PPO (red) and SAC (blue) in the four illustration environments, with 10 different seeds each. 
Snapshots show typical postures of MIMo at the start (top) and end (bottom) of an episode. Videos are available at \url{https://tinyurl.com/MIMo-playlist}}
\label{fig-experiments}
\end{figure*}

\textcolor{black}{We use the default parameters of the Stable-Baselines3 library~\cite{stable-baselines3}}, consisting of linear networks with two hidden layers of size 64 for PPO and size 256 for SAC, \textcolor{black}{as well as common improvements\cite{Engstrom2020Implementation, DBLP:journals/corr/abs-1812-05905} that are critical for performance}. Input and output layer sizes vary depending on the observation and action spaces of the environments. Performance is compared for PPO and SAC with 10 different seeds (Fig.~\ref{fig-experiments}). We do not claim that such extrinsically motivated learning is how human infants learn these skills. We merely use these examples to showcase MIMo learning from multimodal input. 

\paragraph{Reaching for Objects}

Reaching is a complex behavior that emerges in the first 6 months of age.  Since it requires hand-eye coordination, infants must combine vision, proprioception, and touch to produce the desired motion \cite{corbetta2018reaching}. A motor command is generated for the arm and hand muscles to produce the reaching movement towards an object in their visual field, with immediate haptic feedback about its success.

In our illustration, MIMo learns to reach for a ball. He is standing in front of the target, which changes position randomly in each episode, always within reach of MIMo’s right hand. He can only move his right shoulder, elbow, and hand joints. His head and eyes are set to look directly at the ball, i.e., the initial visual search and object fixation is assumed. The observation space only includes the proprioception sensory modality. MIMo can use the joint angles of his head and eyes to determine the position of the target. We introduce an additional difficulty relative to the original version of the task \cite{mattern2022} by also sampling the initial condition of the right arm from 10 random positions. This ensures that MIMo effectively uses his proprioception to reach the ball as fast as possible. The reward function
\begin{equation}
    r = \left\{
        \begin{array}{ c l }
        100 & \textrm{if target reached,} \\
        - \Vert \mathbf{p}_\textrm{fingers} - \mathbf{p}_\textrm{target}\Vert & \textrm{otherwise}
    \end{array}
    \right.
\end{equation}

\noindent is the negative distance between the positions of the fingers and the target, with a sparse positive reward when contact is detected. Each episode lasts 1000 time steps or until MIMo successfully touches the ball.
Results for this task are shown in Fig.~\ref{fig-experiments}a.
\paragraph{Standing up}

Infants learn to stand up by themselves at around 10 months and start walking shortly after. This is a gradual process that includes previous stages such as crawling and maintaining balance. One particular stage is marked by the emergence of pulling-to-stand, when infants who are unable to stand without support grasp the edge of a solid surface and pull themselves upwards, thus combining the strengths of their arms and legs \cite{atun2012stand}. This behavior appears as early as 7 months of age and is a necessary milestone during independent locomotion development.

To reproduce the pulling-to-stand behavior, we design an environment where MIMo is placed sitting inside a crib. His feet are fixed to the ground and his hands are fixed to the crib’s rail guard, at a height of \SI{45}{\centi\metre}. He can move the joints in his arms, torso, and legs, with the aim of standing up. The observation space includes the proprioception and vestibular sensory modalities. The latter can be particularly useful by providing information about vertical acceleration. The extrinsic reward is given by

\begin{equation}
    r = z_\textrm{head} - 0.01 \sum_{i\in \textrm{joints}} u_i^2
\end{equation}

\noindent where \(z_\textrm{head}\) is the head’s height measured from an initial height of \SI{20}{\centi\metre}, \(u_i\) is the control input for joint \(j\), and the sum is taken over all active joints. This reward function favors standing positions while penalizing states that require excessive force. The parameters are set to balance the two components. All episodes last 500 time steps. Results for this task are shown in Fig.~\ref{fig-experiments}b.

\paragraph{Self-body Knowledge}

Infants learn not only about the world but also about themselves and their own bodies. In fact, this begins as a tactile exploratory behavior before birth and continues over the first few months of life \cite{jacquey2020body}. Infants develop a self-body knowledge that allows them to map the multi-modal sensory inputs to the different parts of their bodies.

MIMo can learn this self-body knowledge by using his touch perception. We design an environment where MIMo is sitting with his legs crossed, such that his right arm can reach all of his body parts. In each episode he is given a target body part sampled uniformly at random from the geometric primitives that make up his body. By only moving his right arm, he is trained to activate the touch sensors on the target body part. The observation space includes proprioception and touch, as well as the target as a vector with one-hot encoding. The reward function 

\begin{equation}
    r = \left\{
        \begin{array}{ c l }
        500 & \textrm{if target touched,} \\
        - \Vert \mathbf{p}_\textrm{touched} - \mathbf{p}_\textrm{target}\Vert & \textrm{if other part touched,} \\
        -1 & \textrm{otherwise}
    \end{array}
    \right.
\end{equation}

\noindent is positive only if the target body part is touched. Otherwise, it is either the negative distance to the target body part if another touch signal is activated or a fixed negative value if there is no touch signal. Each episode lasts 500 time steps or until MIMo successfully touches the target. Results for this task are shown in Fig.~\ref{fig-experiments}c.

\paragraph{Catching Objects}

As they develop, infants learn to rely on sensory feedback to adapt their actions in order to achieve their goals. One example is catching a falling object, where visual and haptic information gives a cue about when and how to grasp. While newborns have an innate palmar grasp reflex, they learn to predict the motion of objects to catch them at around 8 months of age \cite{vanhof2008relation}.

We illustrate this behavior in an experiment where MIMo learns to grasp a ball that falls onto his hand. The ball's size, mass, and initial position are randomized in each episode. Using the full hand model, MIMo's body is fixed in a standing position with his right arm stretched in front of him, such that he can only move his right hand wrist and fingers. The thumb is locked to the side. At each time step, his head and eyes are automatically rotated to fixate on the falling ball. His observation space includes proprioception, touch and the size of the ball. MIMo needs to learn to integrate these pieces of information to successfully grasp the ball. This experiment, unlike the others, uses the muscle actuation model. The reward function is given by

\begin{equation}
    r = \left\{
        \begin{array}{ c l }
        -100 & \textrm{if ball falls beyond hand,} \\
        100 & \textrm{if ball is held for 1 second,} \\
        N_{c} - c_w(a) & \textrm{otherwise}
    \end{array}
    \right.
\end{equation}

where \(N_{c}\) is the number of geometric primitives in contact with the ball and \(c_w(a)\) is the cost function for the actuators as given by Equation~\ref{eq-musclecost}. This cost function rewards grasping with the fingers over using the wrist. Each episode lasts 800 time steps, or until either the ball falls beyond the hand's height or MIMo has held onto the ball continuously for a full second.
Results for this task are shown in Fig.~\ref{fig-experiments}d.

\textcolor{black}{Both SAC and PPO quickly converge to the optimal behavior for the simpler tasks, demonstrating that the parametrization of MIMo allows for stable motor control. The two harder tasks, (b) and (d), require either full-body control or difficult hand coordination across all fingers. In these scenarios, SAC performs worse and we observe larger variance across seeds. We conjecture that PPO benefits from its monotonic improvement guarantees and therefore achieves better learning stability and lower variance across seeds.} 

\subsection{Benchmarking MIMo}
\label{sec-benchmarks}

In this section we measure the simulation speed of MIMo. In particular, we are interested in assessing under what conditions we can achieve faster than real-time simulations.

Each benchmark runs for one hour of simulation time. Environments have a maximum episode duration and are reset when a goal is achieved or the time limit is reached. We measure the real-time spent in each run, as well as in each of the different components of the system: MuJoCo and the sensory modalities. Physics steps last \SI{5}{\milli\second} for all benchmarks.  Unless stated otherwise we use the spring-damper actuation model with the mitten hand version of MIMo. Results are reported as real seconds required for each simulation second. The test system is equipped with an AMD FX-8350, 16GB RAM and a GTX 1070. The execution times are measured using Python's cProfile library.

\begin{figure}[tb]
\centering\includegraphics[width=\linewidth]{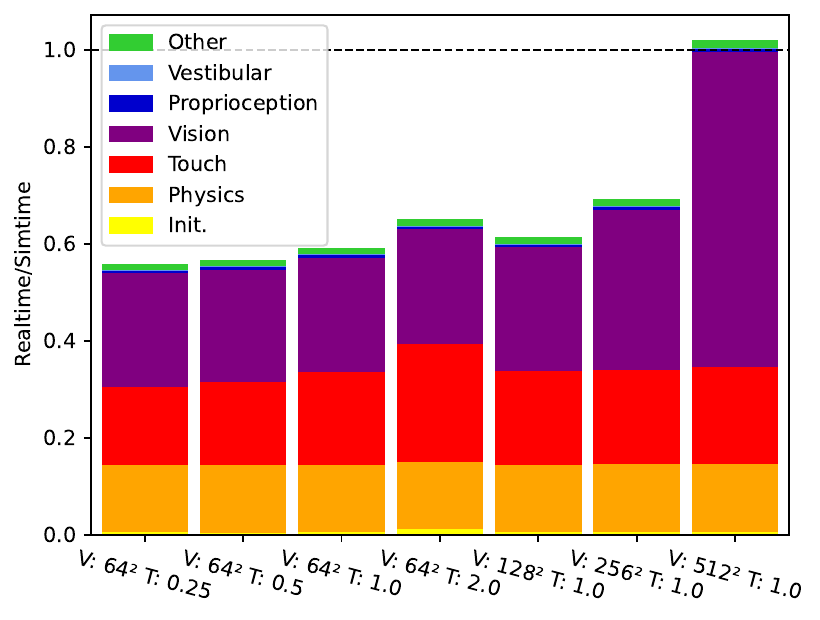}
\caption{Results of the performance benchmarks. Each bar represents one run consisting of 60 episodes of 1-minute length each. The labels indicate the pixel resolution for the visual system (V) and a scalar multiplier for the sensor density for the touch system (T) used. The 4 leftmost bars correspond to configurations with increasing touch sensor density and constant visual resolution, the 3 rightmost bars to increasing visual resolution and constant touch sensor density. All configurations with visual resolutions lower than 512x512 pixels run significantly faster than real-time (dashed horizontal line).}
\label{fig-benchmark-sensors}
\end{figure}

In the first experiment we test performance with multiple configurations for the different sensory modalities, focusing on the vision and touch modules since they are most sensitive to their configuration and consume the bulk of the processing time. The environment consists of MIMo and two objects. MIMo takes random actions continuously. Each episode has a fixed length of 1 minute, with 60 episodes per benchmark. The results can be seen in Fig.~\ref{fig-benchmark-sensors}. The default configuration of MIMo (vision resolution of 256 $ \times $ 256 pixels) \textcolor{black}{requires 0.69 real seconds for each simulation second, i.e., it is 1.44 times faster than real time.}

\begin{figure}[tpb]
\centering
\includegraphics[width=\linewidth]{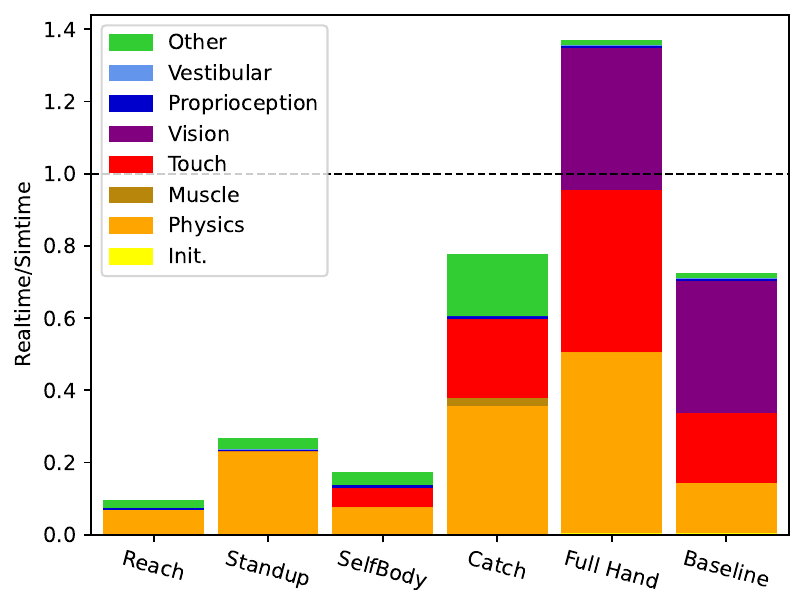}
\caption{Performance benchmarks for each of the demo environments and the full hand version of MIMo. The benchmark of the default configuration, using the mitten-hands, is also plotted for reference.}
\label{fig-benchmark-demo}
\end{figure}

In the next benchmark we test the performance of our demo environments. The number of episodes is no longer fixed and individual episodes may be cut short if MIMo achieves the goal of the environment.
\textcolor{black}{The reach, self-body, and catch experiments perform two physics steps for every control step, the default setting for MIMo environments. The stand-up experiment uses one physics step per control step in order to increase the stability of the simulation due to the additional constraints between MIMo's hands and the crib.}
We also test the performance of the full hand version of MIMo. The configuration is the same as the first benchmark, but with the version of MIMo replaced.
Results for both benchmarks are plotted in Fig.~\ref{fig-benchmark-demo}. The \textcolor{black}{``}Reach", \textcolor{black}{``}Standup", and \textcolor{black}{``}SelfBody" experiment are all based on the default configuration with some modalities disabled and \textcolor{black}{joints locked in place. As a result, they }perform significantly faster. Interestingly, the physics simulation for the stand-up environment takes longer than the baseline. This slowdown comes from:
1. A slowdown in MuJoCo's solver, as MIMo's joint positions are more constrained since both his hands and feet are fixed in place.
2. The environment configuration: MIMo's initial position is slightly randomized and the simulation allowed to settle for several physics steps without any actions before each episode begins.

As to the full hand version, the performance of both MuJoCo and the touch module is highly dependent on the number of contacts in the simulation. Contacts between two body parts of MIMo are particularly expensive as the force distribution has to be computed for both bodies. This leads to a significant slow-down for both MuJoCo and the touch module due to contacts between MIMo's fingers. 
The \textcolor{black}{``}Catch" environment is based on the full hand version, with all joints, except the right hand, locked into fixed positions, and touch sensation only in the right arm and hand. This speeds up both touch and physics performance. Muscle functions make up 3\

The muscle actuation model adds a flat run-time cost of \SI{0.8}{\milli\second} per physics step for both versions of MIMo, which leads to a 23\

Performance can be improved for specific experiments in multiple ways. In addition to adjusting the configuration of the modalities or the scene, the performance of the simulation can be improved by \textcolor{black}{reducing the frequency of control steps}, reducing the frequency of observation collection and thus the time spent in the sensory modules.

\section{Discussion}
\label{sec-discussion}

In this work we have presented MIMo, the multimodal infant model, an open-source research software platform for 1) building models of cognitive development in infants and toddlers and 2) constructing AIs that can learn in a similar self-directed fashion from interactions with their environment. Compared to previous software platforms\cite{tikhanoff2008open,kerzel2017nico,kuniyoshi2006early,tassa2012synthesis,coumans2016pybullet}, a key strength of MIMo is the combination of 1) state-of-the-art physics simulation based on MuJoCo (\url{https://mujoco.org}) with 2) a full-body touch-sensitive skin, and 3) a plausible approximation of muscle-driven movement, while maintaining the ability to run simulations faster than real-time on standard hardware. We believe that these ingredients are essential for advancing our understanding of the development of, e.g., early self-models in infants, that pave the way toward full-fledged adult-like forms of intelligence and consciousness.

In the past, computational models of cognitive development have often been restricted to isolated cognitive phenomena. Some examples are works on
binocular vision \cite{dominguez2003developmental,eckmann2020active},
visual object and category learning \cite{mareschal2000connectionist,schneider2021contrastive,aubret2022toddler},
gaze following \cite{nagai2003constructive,triesch2006gaze},
learning to grasp objects \cite{oztop2004infant},
perservative reaching \cite{thelen2001dynamics},
word learning \cite{roy2002learning,yu2007unified,xu2007word},
and countless others.
While such models have produced many important insights, they often work with simplified sensory inputs and it is not clear how to scale them to the rich multimodal sensory input provided by our sense organs.

Recent approaches that train models with first-person video and audio recordings of infants \cite{bambach2018toddler,orhan2020self} try to overcome this limitation. Critically, however, the model still learns from just {\em passively observing} these inputs recorded in this way, while the actual infant {\em actively generated} this sensory input through their own behavior. Therefore, as we have argued in the Introduction, this first-person-recording approach is likely to miss an essential aspect of cognitive development: the infant's ability to actively probe the causal structure of the world in a targeted fashion. Capturing such more advanced forms of learning in models necessitates modeling environments that support embodied interaction with the physical environment. We feel that the time is ripe to fully embrace such models and we believe that it should be done without incurring the burden and limited reproducability associated with working with humanoid robots.

In the design of MIMo we faced a number of trade-offs between realism and computational efficiency. The ``right'' trade-off will always depend on the particular phenomenon being investigated. The choices we made already permit faster than real-time simulations on today's standard hardware. However, this has resulted in various limitations. For example, MIMo's body is composed of simple rigid shape primitives. Furthermore, MIMo is presently limited to just four sensory modalities (binocular vision, proprioception, touch, and a vestibular system). In the future, we plan to incorporate nociception (pain perception), audition, and possibly olfaction.

\textcolor{black}{Future work could also study the effects of a growing body during development. All relevant aspects, such as MIMo's physical size and strength, can be adjusted even within episodes. A model of the development of, say, a particular sensorimotor skill could be trained while parameters of MIMo's body slowly change.}

We hope that MIMo will facilitate research into how cognition develops from embodied interactions with the physical and social environment, whose rich structure is sensed through multiple modalities. At the very least, it should make such efforts easier and more reproducible. Furthermore, we hope that MIMo will enable a more cumulative and collaborative approach to such research, where models of the development of higher-level cognitive functions are built on top of previously published models of the development of precursor skills. After all, human development is often a cumulative process, where new representations, skills, and competences are built on top of already existing ones in an open-ended fashion. 

\bibliographystyle{IEEEtran}

\bibliography{IEEEabrv,references}

\begin{thebibliography}{10}
\providecommand{\url}[1]{#1}
\csname url@samestyle\endcsname
\providecommand{\newblock}{\relax}
\providecommand{\bibinfo}[2]{#2}
\providecommand{\BIBentrySTDinterwordspacing}{\spaceskip=0pt\relax}
\providecommand{\BIBentryALTinterwordstretchfactor}{4}
\providecommand{\BIBentryALTinterwordspacing}{\spaceskip=\fontdimen2\font plus
\BIBentryALTinterwordstretchfactor\fontdimen3\font minus \fontdimen4\font\relax}
\providecommand{\BIBforeignlanguage}[2]{{%
\expandafter\ifx\csname l@#1\endcsname\relax
\typeout{** WARNING: IEEEtran.bst: No hyphenation pattern has been}%
\typeout{** loaded for the language `#1'. Using the pattern for}%
\typeout{** the default language instead.}%
\else
\language=\csname l@#1\endcsname
\fi
#2}}
\providecommand{\BIBdecl}{\relax}
\BIBdecl

\bibitem{mattern2022}
D.~Mattern, F.~M. L{\'o}pez, M.~R. Ernst, A.~Aubret, and J.~Triesch, ``Mimo: A multi-modal infant model for studying cognitive development in humans and ais,'' in \emph{2022 IEEE International Conference on Development and Learning (ICDL)}.\hskip 1em plus 0.5em minus 0.4em\relax IEEE, 2022, pp. 23--29.

\bibitem{turing1950mind}
A.~M. Turing, ``Computing machinery and intelligence,'' \emph{Mind}, vol.~59, no. 236, pp. 433--460, 1950.

\bibitem{asada2001cognitive}
M.~Asada, K.~F. MacDorman, H.~Ishiguro, and Y.~Kuniyoshi, ``Cognitive developmental robotics as a new paradigm for the design of humanoid robots,'' \emph{Robotics and Autonomous Systems}, vol.~37, no.~2, pp. 185--193, 2001, {H}umanoid {R}obots.

\bibitem{lungarella2003developmental}
M.~Lungarella, G.~Metta, R.~Pfeifer, and G.~Sandini, ``Developmental robotics: a survey,'' \emph{Connection Science}, vol.~15, no.~4, pp. 151--190, 2003.

\bibitem{schmidhuber2006developmental}
J.~Schmidhuber, ``Developmental robotics, optimal artificial curiosity, creativity, music, and the fine arts,'' \emph{Connection Science}, vol.~18, no.~2, pp. 173--187, 2006.

\bibitem{asada2009cognitive}
M.~Asada, K.~Hosoda, Y.~Kuniyoshi, H.~Ishiguro, T.~Inui, Y.~Yoshikawa, M.~Ogino \emph{et~al.}, ``Cognitive developmental robotics: A survey,'' \emph{IEEE Transactions on Autonomous Mental Development}, vol.~1, no.~1, pp. 12--34, 2009.

\bibitem{cangelosi2018babies}
A.~Cangelosi and M.~Schlesinger, ``From babies to robots: The contribution of developmental robotics to developmental psychology,'' \emph{Child Development Perspectives}, vol.~12, no.~3, pp. 183--188, 2018.

\bibitem{doya2019toward}
K.~Doya and T.~Taniguchi, ``Toward evolutionary and developmental intelligence,'' \emph{Current Opinion in Behavioral Sciences}, vol.~29, pp. 91--96, 2019, artificial Intelligence.

\bibitem{gan2020threedworld}
\BIBentryALTinterwordspacing
C.~Gan, J.~Schwartz, S.~Alter, D.~Mrowca, M.~Schrimpf, J.~Traer, J.~D. Freitas \emph{et~al.}, ``Three{DW}orld: A platform for interactive multi-modal physical simulation,'' in \emph{Thirty-fifth Conference on Neural Information Processing Systems Datasets and Benchmarks Track (Round 1)}, 2021. [Online]. Available: \url{https://openreview.net/forum?id=db1InWAwW2T}
\BIBentrySTDinterwordspacing

\bibitem{rohmer2013v}
E.~Rohmer, S.~P.~N. Singh, and M.~Freese, ``V-rep: A versatile and scalable robot simulation framework,'' in \emph{2013 IEEE/RSJ International Conference on Intelligent Robots and Systems}, 2013, pp. 1321--1326.

\bibitem{coumans2016pybullet}
E.~Coumans and Y.~Bai, ``Pybullet, a python module for physics simulation for games, robotics and machine learning,'' 2016.

\bibitem{collins2021review}
J.~Collins, S.~Chand, A.~Vanderkop, and D.~Howard, ``A review of physics simulators for robotic applications,'' \emph{IEEE Access}, vol.~9, pp. 51\,416--51\,431, 2021.

\bibitem{todorov2012mujoco}
E.~Todorov, T.~Erez, and Y.~Tassa, ``Mujoco: A physics engine for model-based control,'' in \emph{2012 IEEE/RSJ International Conference on Intelligent Robots and Systems}, 2012, pp. 5026--5033.

\bibitem{fu20213d}
H.~Fu, R.~Jia, L.~Gao, M.~Gong, B.~Zhao, S.~Maybank, and D.~Tao, ``3d-future: 3d furniture shape with texture,'' \emph{International Journal of Computer Vision}, pp. 1--25, 2021.

\bibitem{Stojanov_2021_CVPR}
S.~Stojanov, A.~Thai, and J.~M. Rehg, ``Using shape to categorize: Low-shot learning with an explicit shape bias,'' in \emph{Proceedings of the IEEE/CVF Conference on Computer Vision and Pattern Recognition (CVPR)}, June 2021, pp. 1798--1808.

\bibitem{olmos2004}
A.~Olmos and F.~A.~A. Kingdom, ``Mcgill calibrated colour image database,'' http://tabby.vision.mcgill.ca.

\bibitem{tikhanoff2008open}
V.~Tikhanoff, A.~Cangelosi, P.~Fitzpatrick, G.~Metta, L.~Natale, and F.~Nori, ``An open-source simulator for cognitive robotics research: The prototype of the i{C}ub humanoid robot simulator,'' in \emph{Proceedings of the 8th Workshop on Performance Metrics for Intelligent Systems}, ser. PerMIS '08.\hskip 1em plus 0.5em minus 0.4em\relax New York, NY, USA: Association for Computing Machinery, 2008, p. 57–61.

\bibitem{kerzel2017nico}
M.~Kerzel, E.~Strahl, S.~Magg, N.~Navarro-Guerrero, S.~Heinrich, and S.~Wermter, ``Nico — neuro-inspired companion: A developmental humanoid robot platform for multimodal interaction,'' in \emph{2017 26th IEEE International Symposium on Robot and Human Interactive Communication (RO-MAN)}, 2017, pp. 113--120.

\bibitem{kuniyoshi2006early}
Y.~Kuniyoshi and S.~Sangawa, ``Early motor development from partially ordered neural-body dynamics: experiments with a cortico-spinal-musculo-skeletal model,'' \emph{Biological Cybernetics}, vol.~95, no.~6, pp. 589--605, 2006.

\bibitem{oztop2004infant}
E.~Oztop, N.~S. Bradley, and M.~A. Arbib, ``Infant grasp learning: a computational model,'' \emph{Experimental brain research}, vol. 158, no.~4, pp. 480--503, 2004.

\bibitem{priamikov2016openeyesim}
A.~Priamikov, M.~Fronius, B.~Shi, and J.~Triesch, ``Openeyesim: a biomechanical model for simulation of closed-loop visual perception,'' \emph{Journal of Vision}, vol.~16, no.~15, pp. 25--25, 12 2016.

\bibitem{delp2007opensim}
S.~L. Delp, F.~C. Anderson, A.~S. Arnold, P.~Loan, A.~Habib, C.~T. John, E.~Guendelman \emph{et~al.}, ``Opensim: Open-source software to create and analyze dynamic simulations of movement,'' \emph{IEEE Transactions on Biomedical Engineering}, vol.~54, no.~11, pp. 1940--1950, 2007.

\bibitem{haarnoja2018sac}
T.~Haarnoja, A.~Zhou, P.~Abbeel, and S.~Levine, ``Soft actor-critic: Off-policy maximum entropy deep reinforcement learning with a stochastic actor,'' in \emph{Proceedings of the 35th International Conference on Machine Learning}, ser. Proceedings of Machine Learning Research, J.~Dy and A.~Krause, Eds., vol.~80.\hskip 1em plus 0.5em minus 0.4em\relax PMLR, 10--15 Jul 2018, pp. 1861--1870.

\bibitem{tassa2012synthesis}
Y.~Tassa, T.~Erez, and E.~Todorov, ``Synthesis and stabilization of complex behaviors through online trajectory optimization,'' in \emph{2012 IEEE/RSJ International Conference on Intelligent Robots and Systems}, 2012, pp. 4906--4913.

\bibitem{anthrokids}
\BIBentryALTinterwordspacing
S.~Ressler, ``Anthrokids-anthropometric data of children,'' \emph{National Institute of Standards and Technology}, 1977. [Online]. Available: \url{https://math.nist.gov/~SRessler/anthrokids/}
\BIBentrySTDinterwordspacing

\bibitem{kumar2016thesis}
\BIBentryALTinterwordspacing
V.~Kumar, ``Manipulators and manipulation in high dimensional spaces,'' Ph.D. dissertation, University of Washington, Seattle, 2016. [Online]. Available: \url{https://digital.lib.washington.edu/researchworks/handle/1773/38104}
\BIBentrySTDinterwordspacing

\bibitem{McKay}
M.~J. McKay, J.~N. Baldwin, P.~Ferreira, M.~Simic, N.~Vanicek, J.~Burns, .~N.~P. Consortium \emph{et~al.}, ``Normative reference values for strength and flexibility of 1,000 children and adults,'' \emph{Neurology}, vol.~88, no.~1, pp. 36--43, 2017.

\bibitem{Eek}
M.~N. Eek, A.-K. Kroksmark, and E.~Beckung, ``Isometric muscle torque in children 5 to 15 years of age: Normative data,'' \emph{Archives of Physical Medicine and Rehabilitation}, vol.~87, no.~8, pp. 1091--1099, 2006.

\bibitem{Sankar}
W.~N. Sankar, C.~T. Laird, and K.~D. Baldwin, ``Hip range of motion in children: what is the norm?'' \emph{Journal of Pediatric Orthopaedics}, vol.~32, no.~4, pp. 399--405, 2012.

\bibitem{Gomez}
T.~Gomez, G.~Beach, C.~Cooke, W.~Hrudey, and P.~Goyert, ``Normative database for trunk range of motion, strength, velocity, and endurance with the isostation b-200 lumbar dynamometer,'' \emph{Spine}, vol.~16, no.~1, p. 15—21, January 1991.

\bibitem{Jordan}
A.~Jordan, J.~Mehlsen, P.~M. B{\"u}low, K.~{\O}stergaard, and B.~Danneskiold-Sams{\o}e, ``Maximal isometric strength of the cervical musculature in 100 healthy volunteers,'' \emph{Spine}, vol.~24, no.~13, p. 1343, 1999.

\bibitem{Ohman}
A.~M. {\"O}hman and E.~R. Beckung, ``Reference values for range of motion and muscle function of the neck in infants,'' \emph{Pediatric Physical Therapy}, vol.~20, no.~1, pp. 53--58, 2008.

\bibitem{Watanabe}
H.~Watanabe, K.~Ogata, T.~Amano, and T.~Okabe, ``The range of joint motions of the extremities in healthy japanese people--the difference according to the age (author's transl),'' \emph{Nihon Seikeigeka Gakkai Zasshi}, vol.~53, no.~3, pp. 275--261, 1979.

\bibitem{Gunal}
I.~G{\"u}nal, N.~K{\"o}se, O.~Erdogan, E.~G{\"o}kt{\"u}rk, and S.~Seber, ``Normal range of motion of the joints of the upper extremity in male subjects, with special reference to side,'' \emph{Journal of Bone \& Joint Surgery}, vol.~78, no.~9, p. 1401, 1996.

\bibitem{Hughes}
R.~E. Hughes, M.~E. Johnson, S.~W. O'Driscoll, and K.-N. An, ``Age-related changes in normal isometric shoulder strength,'' \emph{The American Journal of Sports Medicine}, vol.~27, no.~5, pp. 651--657, 1999.

\bibitem{Katoh}
M.~Katoh, ``Test-retest reliability of isometric shoulder muscle strength measurement with a handheld dynamometer and belt,'' \emph{Journal of Physical Therapy Science}, vol.~27, no.~6, pp. 1719--1722, 2015.

\bibitem{DaPaz}
S.~N. Da~Paz, A.~Stalder, S.~Berger, and K.~Ziebarth, ``Range of motion of the upper extremity in a healthy pediatric population: introduction to normative data,'' \emph{European Journal of Pediatric Surgery}, vol.~26, no.~05, pp. 454--461, 2016.

\bibitem{Bok}
S.-K. Bok, T.~H. Lee, and S.~S. Lee, ``The effects of changes of ankle strength and range of motion according to aging on balance,'' \emph{Annals of Rehabilitation Medicine}, vol.~37, no.~1, pp. 10--16, 2013.

\bibitem{ohtsuki1981decrease}
T.~Ohtsuki, ``Decrease in grip strength induced by simultaneous bilateral exertion with reference to finger strength,'' \emph{Ergonomics}, vol.~24, no.~1, pp. 37--48, 1981.

\bibitem{ekman1992emotions}
P.~Ekman, ``An argument for basic emotions,'' \emph{Cognition and Emotion}, vol.~6, no. 3-4, pp. 169--200, 1992.

\bibitem{vanswearingen1983measuring}
J.~M. Vanswearingen, ``Measuring wrist muscle strength,'' \emph{Journal of Orthopaedic \& Sports Physical Therapy}, vol.~4, no.~4, pp. 217--228, 1983.

\bibitem{wochner2022learning}
\BIBentryALTinterwordspacing
I.~Wochner, P.~Schumacher, G.~Martius, D.~B{\"u}chler, S.~Schmitt, and D.~Haeufle, ``Learning with muscles: Benefits for data-efficiency and robustness in anthropomorphic tasks,'' in \emph{6th Annual Conference on Robot Learning}, 2022. [Online]. Available: \url{https://openreview.net/forum?id=Xo3eOibXCQ8}
\BIBentrySTDinterwordspacing

\bibitem{frey2012knee}
L.~A. Frey-Law, A.~Laake, K.~G. Avin, J.~Heitsman, T.~Marler, and K.~Abdel-Malek, ``Knee and elbow 3d strength surfaces: peak torque-angle-velocity relationships,'' \emph{Journal of applied biomechanics}, vol.~28, no.~6, pp. 726--737, 2012.

\bibitem{dampingmo2020}
\BIBentryALTinterwordspacing
A.~Mo, F.~Izzi, D.~F.~B. Haeufle, and A.~Badri-Spröwitz, ``Effective viscous damping enables morphological computation in legged locomotion,'' \emph{Frontiers in Robotics and AI}, vol.~7, 2020. [Online]. Available: \url{https://www.frontiersin.org/articles/10.3389/frobt.2020.00110}
\BIBentrySTDinterwordspacing

\bibitem{muscledamping22}
\BIBentryALTinterwordspacing
F.~Izzi, A.~Mo, S.~Schmitt, A.~Badri-Spr{\"o}witz, and D.~F.~B. Haeufle, ``Muscle prestimulation tunes velocity preflex in simulated perturbed hopping,'' \emph{Scientific Reports}, vol.~13, no.~1, p. 4559, Mar 2023. [Online]. Available: \url{https://doi.org/10.1038/s41598-023-31179-6}
\BIBentrySTDinterwordspacing

\bibitem{lang1999cerebellar}
C.~E. Lang and A.~J. Bastian, ``Cerebellar subjects show impaired adaptation of anticipatory emg during catching,'' \emph{Journal of neurophysiology}, vol.~82, no.~5, pp. 2108--2119, 1999.

\bibitem{Lee}
W.~J. Lee, J.~H. Kim, Y.~U. Shin, S.~Hwang, and H.~W. Lim, ``Differences in eye movement range based on age and gaze direction,'' \emph{Eye}, vol.~33, no.~7, pp. 1145--1151, 2019.

\bibitem{rosenbaum1999strabismus}
A.~L. Rosenbaum and A.~P. Santiago, \emph{Clinical strabismus management: principles and surgical techniques}.\hskip 1em plus 0.5em minus 0.4em\relax W.B. Saunders, 1999.

\bibitem{strasburger2011vision}
H.~Strasburger, I.~Rentschler, and M.~Jüttner, ``Peripheral vision and pattern recognition: A review,'' \emph{Journal of Vision}, vol.~11, no.~5, pp. 13--13, 12 2011.

\bibitem{johnson1992neural}
K.~O. Johnson and S.~S. Hsiao, ``Neural mechanisms of tactual form and texture perception,'' \emph{Annual Review of Neuroscience}, vol.~15, no.~1, pp. 227--250, 1992.

\bibitem{Mancini}
F.~Mancini, A.~Bauleo, J.~Cole, F.~Lui, C.~A. Porro, P.~Haggard, and G.~D. Iannetti, ``Whole-body mapping of spatial acuity for pain and touch,'' \emph{Annals of Neurology}, vol.~75, no.~6, pp. 917--924, 2014.

\bibitem{brockman2016gym}
G.~Brockman, V.~Cheung, L.~Pettersson, J.~Schneider, J.~Schulman, J.~Tang, and W.~Zaremba, ``{O}pen{AI} {G}ym,'' \emph{arXiv preprint arXiv:1606.01540}, 2016.

\bibitem{stable-baselines3}
A.~Raffin, A.~Hill, A.~Gleave, A.~Kanervisto, M.~Ernestus, and N.~Dormann, ``Stable-baselines3: Reliable reinforcement learning implementations,'' \emph{Journal of Machine Learning Research}, vol.~22, no. 268, pp. 1--8, 2021.

\bibitem{schulman2017ppo}
J.~Schulman, F.~Wolski, P.~Dhariwal, A.~Radford, and O.~Klimov, ``Proximal policy optimization algorithms,'' \emph{arXiv preprint arXiv:1707.06347}, 2017.

\bibitem{pmlr-v37-schulman15}
\BIBentryALTinterwordspacing
J.~Schulman, S.~Levine, P.~Abbeel, M.~Jordan, and P.~Moritz, ``Trust region policy optimization,'' in \emph{Proceedings of the 32nd International Conference on Machine Learning}, ser. Proceedings of Machine Learning Research, F.~Bach and D.~Blei, Eds., vol.~37.\hskip 1em plus 0.5em minus 0.4em\relax Lille, France: PMLR, 07--09 Jul 2015, pp. 1889--1897. [Online]. Available: \url{https://proceedings.mlr.press/v37/schulman15.html}
\BIBentrySTDinterwordspacing

\bibitem{han2021max}
S.~Han and Y.~Sung, ``A max-min entropy framework for reinforcement learning,'' \emph{Advances in Neural Information Processing Systems}, vol.~34, pp. 25\,732--25\,745, 2021.

\bibitem{Engstrom2020Implementation}
\BIBentryALTinterwordspacing
L.~Engstrom, A.~Ilyas, S.~Santurkar, D.~Tsipras, F.~Janoos, L.~Rudolph, and A.~Madry, ``Implementation matters in deep rl: A case study on ppo and trpo,'' in \emph{International Conference on Learning Representations}, 2020. [Online]. Available: \url{https://openreview.net/forum?id=r1etN1rtPB}
\BIBentrySTDinterwordspacing

\bibitem{DBLP:journals/corr/abs-1812-05905}
\BIBentryALTinterwordspacing
T.~Haarnoja, A.~Zhou, K.~Hartikainen, G.~Tucker, S.~Ha, J.~Tan, V.~Kumar \emph{et~al.}, ``Soft actor-critic algorithms and applications,'' \emph{CoRR}, vol. abs/1812.05905, 2018. [Online]. Available: \url{http://arxiv.org/abs/1812.05905}
\BIBentrySTDinterwordspacing

\bibitem{corbetta2018reaching}
D.~Corbetta, R.~F. Wiener, S.~L. Thurman, and E.~G. McMahon, ``The embodied origins of infant reaching: Implications for the emergence of eye-hand coordination,'' \emph{Kinesiology Review}, vol.~7, no.~1, pp. 10--17, 2018.

\bibitem{atun2012stand}
O.~Atun-Einy, S.~E. Berger, and A.~Scher, ``Pulling to stand: Common trajectories and individual differences in development,'' \emph{Developmental Psychobiology}, vol.~54, no.~2, pp. 187--198, 2012.

\bibitem{jacquey2020body}
L.~Jacquey, J.~Fagard, K.~O’Regan, and R.~Esseily, ``Development of body know-how during the baby's first year of life,'' \emph{Enfance}, vol.~2, no.~2, pp. 175--192, 2020.

\bibitem{vanhof2008relation}
P.~van Hof, J.~van~der Kamp, and G.~J. Savelsbergh, ``The relation between infants' perception of catchableness and the control of catching.'' \emph{Developmental Psychology}, vol.~44, no.~1, pp. 182--194, 2008.

\bibitem{dominguez2003developmental}
M.~Dominguez and R.~A. Jacobs, ``Developmental constraints aid the acquisition of binocular disparity sensitivities,'' \emph{Neural Computation}, vol.~15, no.~1, pp. 161--182, 2003.

\bibitem{eckmann2020active}
S.~Eckmann, L.~Klimmasch, B.~E. Shi, and J.~Triesch, ``Active efficient coding explains the development of binocular vision and its failure in amblyopia,'' \emph{Proceedings of the National Academy of Sciences}, vol. 117, no.~11, pp. 6156--6162, 2020.

\bibitem{mareschal2000connectionist}
D.~Mareschal, R.~M. French, and P.~C. Quinn, ``A connectionist account of asymmetric category learning in early infancy.'' \emph{Developmental psychology}, vol.~36, no.~5, p. 635, 2000.

\bibitem{schneider2021contrastive}
F.~Schneider, X.~Xu, M.~R. Ernst, Z.~Yu, and J.~Triesch, ``Contrastive learning through time,'' in \emph{SVRHM 2021 Workshop@ NeurIPS}, 2021.

\bibitem{aubret2022toddler}
A.~Aubret, C.~Teuli{\`e}r, and J.~Triesch, ``Toddler-inspired embodied vision for learning object representations,'' in \emph{2022 IEEE International Conference on Development and Learning (ICDL)}.\hskip 1em plus 0.5em minus 0.4em\relax IEEE, 2022, pp. 81--87.

\bibitem{nagai2003constructive}
Y.~Nagai, K.~Hosoda, A.~Morita, and M.~Asada, ``A constructive model for the development of joint attention,'' \emph{Connection Science}, vol.~15, no.~4, pp. 211--229, 2003.

\bibitem{triesch2006gaze}
J.~Triesch, C.~Teuscher, G.~O. De{\'a}k, and E.~Carlson, ``Gaze following: why (not) learn it?'' \emph{Developmental science}, vol.~9, no.~2, pp. 125--147, 2006.

\bibitem{thelen2001dynamics}
E.~Thelen, G.~Sch{\"o}ner, C.~Scheier, and L.~B. Smith, ``The dynamics of embodiment: A field theory of infant perseverative reaching,'' \emph{Behavioral and brain sciences}, vol.~24, no.~1, pp. 1--34, 2001.

\bibitem{roy2002learning}
D.~K. Roy and A.~P. Pentland, ``Learning words from sights and sounds: A computational model,'' \emph{Cognitive science}, vol.~26, no.~1, pp. 113--146, 2002.

\bibitem{yu2007unified}
C.~Yu and D.~H. Ballard, ``A unified model of early word learning: Integrating statistical and social cues,'' \emph{Neurocomputing}, vol.~70, no. 13-15, pp. 2149--2165, 2007.

\bibitem{xu2007word}
F.~Xu and J.~B. Tenenbaum, ``Word learning as bayesian inference.'' \emph{Psychological review}, vol. 114, no.~2, p. 245, 2007.

\bibitem{bambach2018toddler}
S.~Bambach, D.~Crandall, L.~Smith, and C.~Yu, ``Toddler-inspired visual object learning,'' \emph{Advances in neural information processing systems}, vol.~31, 2018.

\bibitem{orhan2020self}
E.~Orhan, V.~Gupta, and B.~M. Lake, ``Self-supervised learning through the eyes of a child,'' \emph{Advances in Neural Information Processing Systems}, vol.~33, pp. 9960--9971, 2020.

\end{thebibliography}

\end{document}